%% file: cameraready.tex
\pdfoutput=1
\documentclass{article} 
\usepackage[T1]{fontenc}
\usepackage[final]{colm2026_conference}

\usepackage{microtype}
\usepackage{enumitem}
\usepackage{amsmath}
\usepackage{amssymb}
\usepackage{multirow}
\usepackage{pifont}
\usepackage{listings}
\usepackage[most]{tcolorbox}
\usepackage{caption}
\usepackage{subcaption}
\usepackage{graphicx}
\usepackage{float}
\usepackage{tikz}
\usetikzlibrary{positioning,arrows.meta,fit,backgrounds,calc}
\usepackage{amssymb}
\definecolor{c1}{HTML}{0072B2}\definecolor{f1}{HTML}{DEEDF5}
\definecolor{c2}{HTML}{A85585}\definecolor{f2}{HTML}{F4E9EF}
\definecolor{c3}{HTML}{D55E00}\definecolor{f3}{HTML}{FAEADE}
\definecolor{c4}{HTML}{009E73}\definecolor{f4}{HTML}{DEF2ED}
\definecolor{cy}{HTML}{6B4A2F}\definecolor{fy}{HTML}{F3F1EE}
\definecolor{shc}{HTML}{B3261E}
\definecolor{mdl}{HTML}{2F3A45}
\definecolor{ink}{HTML}{1A1A1A}\definecolor{rule}{HTML}{8A8A8A}

\usepackage{booktabs}
\usepackage{array}
\newcolumntype{L}[1]{>{\raggedright\arraybackslash}p{#1}}
\usepackage{inconsolata}
\usepackage{hyperref}
\usepackage{url}
\usepackage{xcolor}

\newcommand{\hlc}[2][yellow!50]{\colorbox{#1}{#2}}
\newcommand{\std}[1]{{\scriptsize$\pm$#1}}
\newcommand{\modelname}{\textsc{UnMask}}

\definecolor{darkblue}{rgb}{0, 0, 0.5}
\hypersetup{colorlinks=true, citecolor=darkblue, linkcolor=darkblue, urlcolor=darkblue}

\title{UNMASK: Discovering and Causally Verifying Spurious Shortcuts in Text Classifiers}

\author{
Chidaksh Ravuru\\
Department of Computer Science\\
University of North Carolina, Chapel Hill\\
\texttt{chidaksh@cs.unc.edu}
\And
Shashank Srivastava\\
Department of Computer Science\\
University of North Carolina, Chapel Hill\\
\texttt{ssrivastava@cs.unc.edu}
}

\begin{document}

\maketitle

\setlength{\textfloatsep}{12pt plus 2pt minus 3pt}
\setlength{\intextsep}{10pt plus 2pt minus 2pt}

\begin{abstract}
Neural language models trained on large crowdsourced corpora frequently exploit spurious surface patterns tied to target labels without true linguistic or causal relevance, boosting benchmark performance while failing on adversarial or out-of-distribution inputs. Existing approaches either require manual specification of the feature vocabulary or automate discovery only partially, leaving the gap between dataset-level correlation and model-level exploitation unaddressed. We present \modelname, a fully automated pipeline that discovers, causally verifies, and mitigates spurious correlations in text classifiers without additional human annotation. Given unlabeled training examples, \modelname{} generates candidate surface patterns as executable boolean expressions, filters them through a statistical validation protocol with independent replication, and establishes causal model dependence via verified counterfactual interventions. Causally confirmed features then serve as annotation-free group definitions for Deep Feature Reweighting, eliminating the group labels that standard DFR requires. Applied to BERT and RoBERTa trained on MNLI, our pipeline independently rediscovers established lexical-overlap and negation biases, verifying 9 of 10 features on BERT and 6 on RoBERTa, and improving HANS accuracy by up to 12.58\,pp. On CivilComments-WILDS, programmatic groups match the 70.1\% worst-group accuracy of hand-labeled DFR~\citep{kirichenko2023layerretrainingsufficientrobustness} without demographic annotation. We further demonstrate that the discovery and validation stages generalize to reward model preference data, surfacing interpretable spurious correlations in RewardBench2.
\end{abstract}

\section{Introduction}

Neural text classifiers can achieve high benchmark accuracy while relying on surface patterns unrelated to the task~\citep{gururangan2018annotationartifactsnaturallanguage,
mccoy-etal-2019-right, gardner2021competencyproblemsfindingremoving}. Significant progress has been made on identifying and mitigating these spurious correlations ~\citep{poliak-etal-2018-hypothesis, wang-etal-2022-identifying, wu2022generatingdatamitigatespurious}, yet each approach requires human intervention at the identification stage.

Automated discovery methods either impose restrictive assumptions or leave the feature vocabulary human-defined. Recent approaches assume access to a bias-free reference corpus~\citep{wang-etal-2022-identifying} or rely on model errors as an opaque proxy for group structure~\citep{liu2021justtraintwiceimproving, creager2021environmentinferenceinvariantlearning}. The fundamental question is: how can we identify which surface features are spurious without knowing them in advance, and once identified, how can we verify that a trained model actually exploits them? For example, the word \emph{never} correlates with contradiction at an odds ratio of 2.86 in MNLI, but this alone does not tell us whether a BERT model trained on MNLI relies on it.

We introduce \modelname{} (Figure~\ref{fig:unmask}), a pipeline that addresses this question at three stages (discovery, causal verification, and mitigation) without additional human annotation. An LLM proposes candidate surface patterns as deterministic boolean functions over the input text. These are deduplicated, validated, and statistically filtered through a two-phase replication protocol (\S\ref{sec:scdata}). For each surviving feature, we generate minimal edits that remove the surface pattern while preserving the semantic label, establishing causal model dependence through paired prediction shifts (\S\ref{sec:causal}). The same boolean expressions then serve as group labels for annotation-free Deep Feature Reweighting in \S\ref{sec:dsc}~\citep{kirichenko2023layerretrainingsufficientrobustness}.

We evaluate \modelname{} on NLI (MNLI, BERT and RoBERTa, six benchmarks) and
CivilComments-WILDS~\citep{koh2021wilds}. Our contributions are:
\textbf{(i)}~A fully automated discovery pipeline that generates and statistically
validates spurious features as executable boolean
expressions~\citep{Ratner_2017}, reusable deterministically across every stage.
\textbf{(ii)}~A counterfactual verification protocol that separates dataset-level
correlation from model-level exploitation. It recovers the established lexical-overlap and
negation biases and reveals a cross-architecture divergence invisible to correlation-only
analysis, in which RoBERTa is immune to contradiction-class features that BERT exploits. On
sentiment it correctly returns zero exploited features, and debiasing on those non-causal
features does not help. \textbf{(iii)}~Annotation-free debiasing via balanced groups mined
with the same expressions, matching hand-labeled
DFR~\citep{kirichenko2023layerretrainingsufficientrobustness} on CivilComments without
demographic annotation. \textbf{(iv)}~Cross-task generalization to CivilComments-WILDS and
RewardBench2 preference data without task-specific modification.

\input{fig_unmask}

\section{Related Work}
\subsection{Spurious Correlations in Text Datasets}
Crowdsourced NLP datasets are known to contain surface-level regularities that correlate with labels but do not reflect the intended task \citep{liu-etal-2022-toward, geva2019modelingtaskannotatorinvestigation}. In NLI, models exploit syntactic heuristics such as lexical overlap and negation bias~\citep{gururangan2018annotationartifactsnaturallanguage, wu2022generatingdatamitigatespurious}. Similar artifacts have been documented in natural language and multimodal tasks such as fact verification \citep{schuster2019debiasingfactverificationmodels} and Visual Question Answering \citep{agarwal2020causalvqarevealingreducing}. \citet{Geirhos_2020} unified these under the shortcut learning framework, and \citet{gardner2021competencyproblemsfindingremoving} demonstrated that systematic biases in the annotation process cause spurious correlations to become more pronounced as datasets scale.

Early mitigation efforts treated the problem as supervised:
\citet{wang-culotta-2020-identifying} train on manually labeled spurious features, and
\citet{wu2022generatingdatamitigatespurious} generate debiased data via LLMs but still
require the target features to be identified upfront.
\citet{wang-etal-2022-identifying} automate discovery through cross-dataset analysis of
model attributions~\citep{sundararajan2017axiomaticattributiondeepnetworks}, which assumes
the pattern is absent from at least one reference corpus, which is rarely true when biases
stem from shared annotation protocols~\citep{gururangan2018annotationartifactsnaturallanguage}.
Other automated approaches stop short of causal verification: in vision,
\citet{zheng2025shortcutprobeprobingpredictionshortcuts} recover latent directions without
interpretable form, and SpurLens~\citep{hosseini2025seeingwhatstherespurious} compares
accuracy with a cue naturally present versus absent, quantifying correlation rather than
causal contribution. \citet{zhou2024explorespuriouscorrelationsconcept} probe concept-level
correlations via concept activation vectors but require predefined concept sets, and
\citet{menon2024discerndecodingsystematicerrors} decode systematic errors into
natural-language descriptions that are not executable. \modelname{} instead produces
boolean expressions that evaluate deterministically and are verified for model-level causal
reliance before debiasing.

Programmatic weak supervision~\citep{Ratner_2017,ratner2018trainingcomplexmodelsmultitask} also uses boolean predicates over text, but its labeling functions are hand-written to \emph{generate} training labels. \modelname{} shares the abstraction while inverting its purpose: expressions are LLM-generated, target spurious rather than label signal, and feed a causal verification stage absent from that paradigm.

\subsection{Counterfactual generation for NLI}
Early approaches to generating contrastive NLI examples relied on
rule-based templates ~\cite{ribeiro-etal-2020-beyond} or manual
annotation \citep{kaushik2020learningdifferencemakesdifference,gardner-etal-2020-evaluating}. While effective for targeted evaluation, template methods are brittle. More recent work such as \citet{wu2022generatingdatamitigatespurious} uses LLMs to automate generation of debiased training data guided by
z-statistics, and \citet{yang2024relationbasedcounterfactualdataaugmentation} generate relation-aware counterfactuals with contrastive learning.
Concurrently, \citet{jaimes2025mitigatingspuriouscorrelationsnli} identify spurious bigrams via log-frequency LMI scores and synthesize label-flipping premise edits, evaluating on SNLI with ELECTRA-Small. \modelname{} differs in that features are deterministic boolean expressions rather than n-gram statistics and are therefore reusable as group labels. Top-$k$ ranking is replaced by a two-stage protocol with FDR control and held-out replication. A causal verification stage filters out high-OR features the model does not in fact rely on.

\subsection{Debiasing Spurious Correlations}
Given identified biases, several families of mitigation strategies
have been proposed. Product-of-Experts methods train an auxiliary
bias-only model and down-weight examples it predicts
confidently~\citep{clark2019donteasywayout, karimi-mahabadi-etal-2020-end,sanh2020learningothersmistakesavoiding}. \citet{nam2020learningfailuretrainingdebiased} train a bias-amplified model and upweight its failures, removing the need for explicit bias specification.
Group DRO~\citep{sagawa2020distributionallyrobustneuralnetworks} directly minimizes worst-group loss, and annotation-free variants infer groups from model errors~\citep{liu2021justtraintwiceimproving} or retrain only the last classification
layer~\citep{kirichenko2023layerretrainingsufficientrobustness}.
\citet{sohoni2022subclassleftbehindfinegrained} discover groups by clustering ERM feature representations with UMAP and GMM before applying Group DRO, but the resulting cluster labels are opaque numeric indices. \modelname{} instead produces boolean expressions that are human-readable and reusable as group labels, counterfactual templates, and debiasing inputs across all pipeline stages.

\section{Methodology}

\subsection{Problem Setup}
\label{sec:ps}
Let $\mathcal{D}_{\mathrm{tr}} = \{(x_i, y_i)\}_{i=1}^{N} \sim P_{\mathrm{tr}}(X,Y)$ denote the training data, where $x_i \in \mathcal{X}$ and $y_i \in \mathcal{Y} := \{1,\dots,K\}$. Let $A$ denote an attribute variable (not necessarily observed) that may be spuriously correlated with $Y$ under $P_{\mathrm{tr}}$. We define a spurious correlation or data artifact as a label-attribute pair $(y,a)$ satisfying $P_{\mathrm{tr}}(Y=y \mid A=a) \neq P_{\mathrm{tr}}(Y=y)$. Let $\mathcal{S} \subseteq \mathcal{Y} \times \mathcal{A}$ denote the set of such pairs, and define the corresponding set of spuriously correlated attributes as $\mathcal{A}_{\mathrm{sub}} := \{ a \in \mathcal{A} : \exists\, y \in \mathcal{Y} \text{ s.t.\ } (y,a) \in \mathcal{S} \}$. We consider a classifier $f_\theta : \mathcal{X} \to \Delta^{K-1}$ trained via empirical risk minimization, $\theta^\star = \arg\min_{\theta} \frac{1}{N} \sum_{i=1}^{N} \ell(f_\theta(x_i), y_i)$, where $\ell$ is a standard classification loss. Among $\mathcal{A}_{\mathrm{sub}}$, the trained model may rely only on a subset $\mathcal{A}' \subseteq \mathcal{A}_{\mathrm{sub}}$, the attributes whose induced correlations are encoded and exploited by $f_\theta$. Our objective is two-fold: data-level artifact discovery followed by causal analysis of model-level reliance.

\subsection{Spurious Correlations in the Dataset}
\label{sec:scdata}
\subsubsection{Candidate Spurious Generation}
\label{sec:scg}
We construct an initial candidate set using an LLM-based generator, \textsc{SCGenLLM}, which proposes candidate patterns and is distinct from the \textsc{GeneratorLLM} that later produces counterfactual edits (\S\ref{sec:causal}). Given training samples $\{x_i\}_{i=1}^{N}$ without labels,\footnote{By withholding labels, we ensure that candidate features are generated without any label-specific information.} we prompt \textsc{SCGenLLM} to propose surface-level patterns that appear consistently across examples but are not logically necessary for the classification task. For each candidate, \textsc{SCGenLLM} returns:
(i)~a natural-language pattern description,
(ii)~a category tag (lexical, structural, or relational) \footnote{The category tag is not used in any of the downstream stages. Preliminary experiments showed that including it improved the diversity of generated patterns.},
(iii)~a brief justification, and
(iv)~an executable boolean expression $b(x)$ that deterministically evaluates to \textsc{true}/\textsc{false} on any input~$x$.
The boolean expression is the key design choice here, enabling fully deterministic evaluation in all downstream stages (Section~\ref{sec:sfv}). To mitigate long-context hallucination, we partition the data into mini-batches and query the LLM independently per batch, aggregating outputs into the raw candidate set $\mathcal{SC}_{\mathrm{cand}}$.

\subsubsection{Initial Deduplication}
\label{sec:dedup}
The raw set $\mathcal{SC}_{\mathrm{cand}}$ contains substantial redundancy. We apply a first deduplication pass that operates on descriptions alone, without yet relying on boolean expression correctness. First, we remove exact duplicates by normalizing and hashing each candidate's boolean expression and pattern description. Second, we compute sentence embeddings over the pattern description only and cluster candidates via a greedy cosine-similarity procedure with threshold~$\tau_1$, yielding $\mathcal{SC}_{\mathrm{init}}$ -- a reduced set free of exact and near-description duplicates.

\subsubsection{Boolean Logic Validation}
\label{sec:blv}

LLM-generated boolean expressions may contain syntax errors, semantic mismatches, or edge-case failures. Since all downstream evaluation depends on expression fidelity, we verify and repair each candidate before it enters the statistical pipeline. For each pattern in $\mathcal{SC}_{\mathrm{init}}$, we first execute its boolean expression on the original source samples to produce an execution trace (per-sample outputs and runtime errors). An \textsc{EvaluatorLLM} then receives the pattern description, boolean expression, and execution trace, and judges correctness. If incorrect, the \textsc{SCGenLLM} rewrites the expression using the evaluator's feedback, the pattern context, and the source samples. This evaluate-rewrite cycle repeats for $n_{\mathrm{iter}}$ rounds. A final \texttt{compile()} gate rejects any expression that cannot be parsed as a valid single-line Python expression. Only confirmed-correct patterns survive, yielding $\mathcal{SC}_{\mathrm{valid}}$.

\subsubsection{Coverage Based Deduplication}
\label{sec:cov_dedup}
With boolean expressions now validated, we perform a second deduplication pass that exploits actual coverage information. For each pattern in $\mathcal{SC}_{\mathrm{valid}}$, we compute its coverage set -- the indices of training samples on which its expression evaluates to \textsc{true}. We then embed each candidate by concatenating its boolean expression and pattern description. Clustering follows the same greedy procedure as before, but with a lower similarity threshold~$\tau_2$. An additional requirement that two candidates are merged only if the Jaccard similarity of their coverage sets also exceeds a minimum overlap threshold is introduced. This joint criterion ensures that patterns firing on largely disjoint subsets of the data are not collapsed even when their surface descriptions are similar. Within each cluster, the candidate with the highest coverage is retained, yielding $\mathcal{SC}_{\mathrm{dist}}$.

\subsubsection{Spurious Feature Validation}
\label{sec:sfv}

The deduplicated set $\mathcal{SC}_{\mathrm{dist}}$ may still contain patterns that are statistically coincidental, definitional, or otherwise non-spurious. We apply a two-phase {Spurious Feature Validation} (SFV) protocol with independent replication.
\paragraph{Feature Evaluation}
Each boolean expression $b_i \in \mathcal{SC}_{\mathrm{dist}}$ is executed on every sample in $\mathcal{D}_{\mathrm{tr}}$, yielding a binary annotation matrix $\mathbf{M} \in \{0,1\}^{N \times |\mathcal{SC}_{\mathrm{dist}}|}$ where $m_{ji} = b_i(x_j)$. We then partition the row indices via label-stratified splitting into disjoint discovery and validation sets ($\mathcal{I}_{\mathrm{disc}}$, $\mathcal{I}_{\mathrm{val}}$), producing sub-matrices $\mathbf{M}_{\mathrm{disc}}$, $\mathbf{M}_{\mathrm{val}}$ with identical column semantics.
 
\paragraph{Discovery Phase}
For each pair $(b_i, y_k)$ we build a $2 \times 2$ contingency table from
$\mathbf{M}_{\mathrm{disc}}$, apply Fisher's exact test \cite{10.2307/2982890}, and control
the false discovery rate at level $\alpha$ via Benjamini-Hochberg
\cite{888cd474-50a6-33fd-a789-415b80e67e78} across all
$|\mathcal{SC}_{\mathrm{dist}}| \times K$ pairs. A pair passes discovery if its BH-adjusted
$p$-value is below $\alpha$, its odds ratio $\mathrm{OR}(b_i, y_k) \geq \mathrm{OR}_{\min}$,
and the feature fires on at least $n_{\min}$ samples. For surviving pairs we also compute
$\mathrm{Coverage}(b_i, y_k) = P(b_i(x){=}1 \mid y{=}y_k)$ and
$\mathrm{Precision}(b_i, y_k) = P(y{=}y_k \mid b_i(x){=}1)$ over $\mathcal{I}_{\mathrm{disc}}$.

\paragraph{Validation, Filtering, and Selection}
Discovery-significant pairs are re-tested on $\mathbf{M}_{\mathrm{val}}$ and replicate only
if they remain significant with $\mathrm{OR} \geq \mathrm{OR}_{\min}$ on held-out data. Two
post-hoc filters then apply: a coverage ceiling
($\min(\text{Cov}_{\mathrm{disc}}, \text{Cov}_{\mathrm{val}}) > \gamma_{\mathrm{cov}}$,
excluded as definitional) and a precision floor
($\max(\text{Prec}_{\mathrm{disc}}, \text{Prec}_{\mathrm{val}}) < \gamma_{\mathrm{prec}}$,
excluded as non-predictive).\footnote{Both are optional, but allow definitional features to be excluded on highly biased subsets.}
Survivors are scored by a weighted average of log odds, coverage, and precision. To remove
residual redundancy we cluster by co-occurrence correlation in $\mathbf{M}_{\mathrm{val}}$
(hierarchical agglomerative \cite{mullner2011modernhierarchicalagglomerativeclustering},
average linkage), take each cluster's highest-scoring representative, and return the
top-$k$ as $\mathcal{F}_{\mathrm{val}}$. The full flow is
$\mathcal{SC}_{\mathrm{cand}} \to \mathcal{SC}_{\mathrm{init}} \to \mathcal{SC}_{\mathrm{valid}} \to \mathcal{SC}_{\mathrm{dist}} \to \mathcal{F}_{\mathrm{val}}$.

\subsection{Spurious Correlation Exploitation}
\label{sec:sce}
\subsubsection{Spurious Reliance Screening}
\label{sec:reliance_screen}
Before investing in counterfactual generation, we screen each feature for
\emph{spurious reliance} on real training data. For each $b_i \in \mathcal{F}_{\mathrm{val}}$
with associated label $y_k$ we evaluate $b_i(x)$ over $\mathcal{D}_{\mathrm{tr}}$ and
partition examples on feature presence $\times$ label agreement. The discriminative cell is
$(b_i(x){=}1,\, y{\neq}y_k)$, where the surface cue is present but the correct label
contradicts it. A model relying on $b_i$ will still predict $y_k$ there at an elevated rate.

We therefore compare the associated-label prediction rate on that counter-evidence group,
$r_{\mathrm{present}}(b_i) = P(\hat{y} = y_k \mid b_i(x){=}1,\; y{\neq}y_k)$, against the
feature-absent control rate $r_{\mathrm{absent}}(b_i)$ via a one-sided two-proportion
$z$-test ($H_1\colon r_{\mathrm{present}} > r_{\mathrm{absent}}$). Features significant at
$\alpha_{\mathrm{suf}}$ pass. The rest are excluded before counterfactual generation,
yielding $\mathcal{F}_{\mathrm{suf}} \subseteq \mathcal{F}_{\mathrm{val}}$.

\subsubsection{Causal Verification}
\label{sec:causal}
Spurious reliance screening (\S\ref{sec:reliance_screen}) provides behavioral evidence that $f_\theta$ over-predicts $y_k$ when $b_i$ is present. However, it does not establish causality. The elevated prediction rate could reflect confounds rather than direct feature exploitation. We establish causality through counterfactual interventions on $b_i$.

\paragraph{Counterfactual Generation}
For each $b_i \in \mathcal{F}_{\mathrm{suf}}$ with associated label $y_k$, we sample from the feature-present pool $\mathcal{P}_i^{+} = \{x_j : b_i(x_j) = 1\}$. A \textsc{GeneratorLLM} produces a minimal edit $x_j^{\mathrm{cf}}$ such that $b_i(x_j^{\mathrm{cf}}) = 0$ while the semantic relationship of the NLI pair is preserved. Feature removal is verified programmatically via $b_i$ and semantic preservation is confirmed by an independent \textsc{EvaluatorLLM}. Rejected candidates receive targeted feedback appended to a cumulative attempt history, which the generator receives in full at each subsequent iteration to avoid previously failed strategies. The loop continues until acceptance or $T_{\max}$ iterations. When several verified features fire on one input the generator targets them jointly, the default on both tasks, and $b_i$ is credited only when its own expression is verified removed. On CivilComments a stricter single-target variant, removing $b_i$ while preserving the others, yields the Tier~A subset (\S\ref{sec:sc_mnli}). Appendix~\ref{app:cf_example} contains a qualitative counterfactual example.  
\paragraph{Effect Measurement and Causal Classification}
For each accepted pair $(x_j, x_j^{\mathrm{cf}})$, we compute the paired shift in the model's probability for the spurious label: $\Delta p_{j} = p_{\theta}(y_k \mid x_j^{\mathrm{cf}}) - p_{\theta}(y_k \mid x_j)$ 
We test whether $\bar{\Delta p}$ is significantly negative via a paired $t$-test, reporting 95\% confidence intervals and Cohen's $d$. A feature $b_i$ is classified as causally exploited if three conditions jointly hold: (i)~$\bar{\Delta p} < - \varepsilon$ (removing the feature decreases $f_\theta$'s predicted probability for $y_k$) (ii)~the effect is statistically significant at level $\alpha_{\mathrm{causal}}$ and (iii)~the direction is correct. Features where removal unexpectedly increases the spurious label probability are flagged and excluded regardless of significance. The resulting set $\mathcal{F}_{\mathrm{causal}} \subseteq \mathcal{F}_{\mathrm{suf}}$ constitutes the final collection of spurious correlations that are statistically validated, model-dependent, and causally exploited by~$f_\theta$.

\subsection{Debiasing Spurious Correlations}
\label{sec:dsc}
The same boolean expressions that drive causal verification serve a final role in programmatically constructing balanced groups for debiasing, without any human annotation. For each $b_i \in \mathcal{F}_{\mathrm{causal}}$ and a $K$-class label space, we instantiate the $2K$ groups defined by the Cartesian product $\{b_i(x){=}1, b_i(x){=}0\} \times \{y_1, \ldots, y_K\}$, generalizing the
presence-by-label-agreement partition of \S\ref{sec:reliance_screen}. Groups are
subsampled to equal size per feature and pooled across all features in
$\mathcal{F}_{\mathrm{causal}}$ with deduplication. This partition serves as the input to the group-based methods evaluated in \S\ref{sec:dfr_approaches}, each of which operationalizes it differently. SCER on NLI instead uses the canonical negation indicator, and JTT uses no group labels at all.

\section{Experiments}
\label{sec:exps}

\subsection{Datasets, Models, and Metrics}
\label{sec:datasets}
\label{sec:metrics}

We evaluate on three task settings. For \textbf{natural language inference} we train on
MNLI~\citep{williams2018broadcoveragechallengecorpussentence} and evaluate on MNLI-matched
and -mismatched, SNLI~\citep{bowman-etal-2015-large},
ANLI R1--R3~\citep{nie-etal-2020-adversarial}, and HANS~\citep{mccoy-etal-2019-right}.
For \textbf{toxicity detection} we use CivilComments-WILDS~\citep{koh2021wilds}
($\approx$269k training examples, 8 annotated demographic identities), reporting 16-cell
worst-group accuracy (WGA: 2 labels $\times$ 8 identities) on the WILDS test split.
For \textbf{sentiment classification} we use SST-2 and IMDB as a controlled ablation
(\S\ref{sec:sentiment_ablation}). We additionally apply the discovery and validation stages
(\S\ref{sec:scdata}) to \textbf{RewardBench2}~\citep{malik2025rewardbench2advancingreward}
as a qualitative case study, showing that the pipeline transfers to reward-model preference
data without task-specific modification.

We train BERT-base-uncased~\citep{devlin2019bertpretrainingdeepbidirectional} and
RoBERTa-base\footnote{These models were chosen because \citet{wu2022generatingdatamitigatespurious} established which specific spurious features each exploits, providing a near-ground-truth reference for pipeline validation.}~\citep{liu2019robertarobustlyoptimizedbert}
via standard empirical risk minimization, and sample $N{=}5{,}000$ balanced instances from
each task's training data for discovery. GPT-4o~\citep{openai2024gpt4ocard} serves as
\textsc{GeneratorLLM} and Qwen3-32B~\citep{yang2025qwen3technicalreport} as
\textsc{EvaluatorLLM}. Decoupling the two across model families reduces confirmation bias
and preference leakage~\citep{panickssery2024llmevaluatorsrecognizefavor,li2026preferenceleakagecontaminationproblem}.
The \textsc{EvaluatorLLM} choice is validated against
ChaosNLI~\citep{nie2020chaosnli}, which provides multiple human annotations per example
rather than a single gold label (Appendix~\ref{app:model_selection}).

For discovery and causal verification we report odds ratio, coverage, precision, and
$\bar{\Delta p}$ with paired $t$-test significance (Appendix
Table~\ref{tab:spurious_features}). Sagawa-WGA, proportional $|\bar{\Delta p}|$ reduction, and
Tier~A vs.\ Tier~C are in Appendices~\ref{app:sagawa_wga},~\ref{app:causal_effectiveness},
and~\ref{app:cc_full}.

\subsection{Debiasing Methods}
\label{sec:dfr_approaches}

Each method consumes $\mathcal{F}_{\mathrm{causal}}$ differently. \textbf{DFR}~\citep{kirichenko2023layerretrainingsufficientrobustness} retrains only the classification head via $\ell_2$-regularized logistic regression on programmatically balanced groups (encoder frozen). \textbf{DFR-IID} augments those groups with a random i.i.d.\ subsample of the training data to preserve in-distribution coverage. \textbf{DFR-FT} extends DFR to full model fine-tuning. \textbf{PoE}~\citep{clark2019donteasywayout,karimi-mahabadi-etal-2020-end} combines a shallow bias-only model over binary feature vectors with the main model. \textbf{PoE-IPW-Group} augments PoE with per-feature inverse-probability weighting across label $\times$ feature-presence groups. \textbf{SCER}~\citep{park2026spuriouscorrelationawareembeddingregularization} applies direction regularization with worst-group-error loss, using the canonical negation indicator~\citep{sagawa2020distributionallyrobustneuralnetworks} for NLI and the disjunction of all SFV features for CivilComments. \textbf{JTT}~\citep{liu2021justtraintwiceimproving} upweights ERM misclassifications, and is excluded from CivilComments where the BERT ERM error rate (${\approx}3\%$) is too low to converge. \textbf{LEACE}~\citep{belrose2023leace} erases causally verified feature directions via closed-form orthogonal projection. Additional DFR variants are in Appendix~\ref{app:causal_effectiveness} and~\ref{app:l1_audit}.

\subsection{Experimental Setup}
\label{sec:exp_setup}

NLI and CivilComments experiments use three random seeds (42, 123, 323) on both
architectures, with the top-$k$ SFV features held constant so that only the
model-dependent stages (\S\ref{sec:sce}) and debiasing runs vary. All experiments run on
$2{\times}$ NVIDIA RTX A6000 GPUs.
Appendix~\ref{app:model_selection} reports ablations validating programmatic over
LLM-based feature evaluation, and Appendix~\ref{app:hyperparams} gives full prompts and
hyperparameters.

\section{Results}
\label{sec:results}

\subsection{Pipeline Discovery and Validation}
\label{sec:sc_mnli}

The SFV stage selects 10 features from MNLI (Appendix Table~\ref{tab:spurious_features}). The pipeline recovers known annotation artifacts: three lexical-overlap variants for entailment, negation-in-hypothesis and contradictory absolutes for contradiction~\citep{gururangan2018annotationartifactsnaturallanguage, mccoy-etal-2019-right}, and a hypothesis-length pattern for neutral. Causal verification confirms 9 of the 10 on BERT and 6 on RoBERTa, consistent across all 3 seeds.\footnote{pp denotes percentage points throughout. A $\bar{\Delta p}$ of $-$15 indicates a 15\,pp drop in the model's predicted probability for the spurious label after counterfactual feature removal.}

The cross-architecture divergence is itself a substantive finding. RoBERTa is immune to
three contradiction-class features that BERT reliably exploits, with effect sizes below
$\varepsilon{=}0.03$ and 95\% confidence intervals crossing zero on all three seeds. One
feature (always/every in premise \& never/no in hypothesis, OR\,=\,10.01) fails on
\emph{both} architectures, so a high dataset-level odds ratio does not guarantee
model-level exploitation~\citep{srikanth-rudinger-2022-partial}. Per-feature sample sizes
and $t$-statistics are in Appendix~\ref{app:cf_stats}.

On CivilComments-WILDS, SFV mines 10 features. Six of them contain WILDS demographic
identity tokens and together recover six of the eight canonical identities, leaving
\emph{male} and \emph{female} unrecovered. A seventh is thematically related but token-free, and the remaining three are absent from that label
set entirely (derogatory language, interjections, second-person pronoun frequency). Causal verification confirms all 10 as jointly
exploited by both architectures. Under the stricter single-target test one feature falls
short on BERT, giving the 9-feature Tier~A subset used below (Appendix
Table~\ref{tab:cc_spurious_features}).
The full pipeline funnel is reported in Appendix Table~\ref{tab:pipeline_funnel}.

\subsection{Debiasing Results}
\label{sec:main_results}

\paragraph{NLI benchmark accuracy.}
Table~\ref{tab:benchmark_acc} reports accuracy across all methods and both architectures.
On BERT, PoE-IPW-Group gives the strongest HANS improvement (64.99\std{4.44},
$+$12.58\,pp over ERM), with PoE close behind, both preserving MNLI-m within 1.2\,pp of
ERM. DFR-FT retains the best in-distribution accuracy with competitive HANS. LEACE and
DFR head-only are ineffective on BERT: the former costs ${\approx}$15\,pp of MNLI-m, the
latter costs 3.4\,pp without a compensating robustness gain. On RoBERTa, ERM already reaches 74.48\std{1.44} on HANS,
leaving less headroom, though PoE-IPW-Group still gains $+$4.08\,pp.
Sagawa worst-group accuracy is secondary here (Appendix~\ref{app:sagawa_wga}) because its
negation partition overlaps the debiasing target features, a confound HANS avoids.

\begin{table}[t!]
\centering
\footnotesize
\setlength{\tabcolsep}{3pt}
\resizebox{\textwidth}{!}{%
\begin{tabular}{llccccccc}
\toprule
 & & \multicolumn{2}{c}{\textbf{MNLI}} & & \multicolumn{3}{c}{\textbf{ANLI}} & \\
\cmidrule(lr){3-4}\cmidrule(lr){6-8}
\textbf{Model} & \textbf{Method} & \textbf{m} & \textbf{mm} & \textbf{SNLI} & \textbf{R1} & \textbf{R2} & \textbf{R3} & \textbf{HANS} \\
\midrule
\multirow{7}{*}{BERT}
 & ERM           & 84.61\std{0.19} & 85.03\std{0.42} & 80.09\std{0.35} & 23.57\std{1.76} & 27.87\std{0.59} & 29.89\std{0.25} & 52.41\std{1.65} \\
 & DFR-FT        & \textbf{83.85\std{0.07}} & \textbf{84.25\std{0.39}} & 79.57\std{0.82} & 24.50\std{0.78} & 27.97\std{1.33} & 30.69\std{0.35} & 56.84\std{3.54} \\
 & PoE           & 83.72\std{0.31} & 83.96\std{0.42} & 80.24\std{0.52} & 22.87\std{0.61} & 28.40\std{0.66} & 29.94\std{0.65} & 63.84\std{4.08} \\
 & PoE-IPW-Group & 83.43\std{0.37} & 83.74\std{0.47} & 80.04\std{0.46} & 23.23\std{0.58} & 27.93\std{0.12} & 30.28\std{0.83} & \textbf{64.99\std{4.44}} \\
 & SCER          & 83.60\std{0.15} & 84.12\std{0.21} & \textbf{80.34\std{0.56}} & 23.17\std{0.86} & 28.33\std{0.55} & 29.97\std{1.32} & 59.59\std{2.87} \\
 & LEACE         & 69.08\std{3.49} & 69.30\std{3.80} & 62.56\std{3.61} & \textbf{29.70\std{0.82}} & \textbf{32.57\std{2.11}} & \textbf{32.92\std{1.80}} & 57.34\std{4.19} \\
 & JTT           & 80.19\std{0.31} & 79.86\std{0.83} & 75.43\std{0.86} & 24.17\std{2.30} & 29.60\std{1.18} & 31.56\std{0.83} & 58.22\std{2.06} \\
\midrule
\multirow{7}{*}{RoBERTa}
 & ERM           & 87.56\std{0.21} & 87.36\std{0.24} & 84.26\std{1.10} & 32.07\std{0.49} & 27.00\std{1.60} & 29.86\std{1.14} & 74.48\std{1.44} \\
 & DFR-FT        & 87.28\std{0.43} & \textbf{87.07\std{0.12}} & 84.24\std{0.61} & 32.73\std{1.33} & 27.90\std{2.39} & 29.92\std{0.79} & 76.73\std{0.73} \\
 & PoE           & 87.25\std{0.11} & 86.83\std{0.09} & 84.68\std{0.33} & 31.80\std{0.89} & 27.37\std{0.93} & 29.81\std{0.88} & 78.01\std{1.14} \\
 & PoE-IPW-Group & 87.01\std{0.04} & 86.68\std{0.10} & 84.42\std{0.43} & 32.57\std{1.65} & 26.50\std{0.70} & 29.47\std{0.67} & \textbf{78.56\std{0.63}} \\
 & SCER          & \textbf{87.29\std{0.10}} & 86.87\std{0.07} & \textbf{84.73\std{0.42}} & 31.03\std{0.57} & 27.63\std{0.90} & 29.08\std{0.60} & 75.29\std{1.17} \\
 & LEACE         & 83.65\std{0.43} & 83.52\std{0.52} & 82.25\std{1.09} & \textbf{33.93\std{1.19}} & \textbf{29.07\std{0.67}} & \textbf{30.19\std{1.51}} & 71.57\std{5.44} \\
 & JTT           & 84.02\std{0.16} & 84.25\std{0.30} & 81.85\std{0.32} & 31.33\std{0.31} & 27.27\std{0.68} & 28.94\std{0.38} & 72.45\std{0.32} \\
\bottomrule
\end{tabular}}
\caption{NLI benchmark accuracy (\%) across all six evaluation sets and HANS. 3-seed mean\std{std}. DFR head-only is omitted for space: it tracks ERM on RoBERTa but loses 3.4\,pp of MNLI-m on BERT, where its worst-group accuracy also collapses (Appendix~\ref{app:sagawa_wga}). \textbf{Bold} = best debiasing result per column per architecture. HANS subcategory breakdowns are in Appendix~\ref{app:hans_breakdown}.}
\label{tab:benchmark_acc}
\end{table}

\paragraph{Method ranking reversal.}
No single method dominates both tasks. PoE-IPW-Group leads on NLI HANS yet trails DFR on
CivilComments WGA, and DFR wins on CivilComments yet is no better than ERM on NLI, where a
frozen head cannot reweight syntactic patterns spread across three label classes.
Per-method causal effectiveness, the cross-architecture analysis, and zero-shot transfer
are in Appendices~\ref{app:causal_effectiveness},~\ref{app:cross_arch}, and~\ref{app:zero_shot}.

\paragraph{CivilComments worst-group accuracy.}
DFR wins on CivilComments for both architectures (Table~\ref{tab:cc_wga}): BERT reaches
71.84\std{1.94} WGA with Tier~A and RoBERTa 72.12\std{0.26} with Tier~C, matching
\citet{kirichenko2023layerretrainingsufficientrobustness}'s 70.1\% for hand-labeled DFR and
showing that programmatic groups substitute for manual annotation. SCER and PoE fall well
below DFR here (67.06\std{0.49} and 65.96\std{0.99} on BERT), reversing the NLI ranking.
RoBERTa's ERM WGA (55.66\std{0.67}) is 3.31\,pp \emph{below} BERT's on CivilComments, the
opposite direction from NLI: stronger pretraining defeats lexical-overlap heuristics but
does not decorrelate demographic identity from toxicity.

\begin{table}[t!]
\centering
\begin{subtable}[t]{0.53\linewidth}
\centering\footnotesize\setlength{\tabcolsep}{3pt}
\begin{tabular}{lcc}
\toprule
\textbf{Method} & \textbf{BERT} & \textbf{RoBERTa} \\
\midrule
ERM                    & 58.97\std{1.32} & 55.66\std{0.67} \\
\textbf{DFR}           & \textbf{71.84\std{1.94}} & \textbf{72.12\std{0.26}} \\
DFR-FT                 & 67.03\std{2.73} & 71.78\std{1.71} \\
SCER                   & 67.06\std{0.49} & 67.57\std{0.89} \\
PoE                    & 65.96\std{0.99} & 64.88\std{1.58} \\
PoE-IPW-Group          & 64.66\std{1.03} & 61.81\std{0.34} \\
LEACE                  & 65.41\std{0.93} & 64.55\std{0.89} \\
Kirichenko '23~(lit.)  & 70.1\phantom{\std{0.00}} & - \\
\bottomrule
\end{tabular}
\subcaption{Debiasing methods (WGA \%).}
\label{tab:cc_wga}
\end{subtable}\hfill
\begin{subtable}[t]{0.45\linewidth}
\centering\footnotesize\setlength{\tabcolsep}{3pt}
\begin{tabular}{lcc}
\toprule
\textbf{Method} & \textbf{WGA} & \textbf{avg-acc} \\
\midrule
ERM        & 58.97\std{1.32} & 92.59\std{0.02} \\
PMI        & 64.80\std{6.91} & 90.44\std{1.04} \\
PMI+SFV    & 68.19\std{0.85} & 90.51\std{0.15} \\
LLM-only   & 65.10\std{4.21} & 89.71\std{0.53} \\
\midrule
\textbf{\modelname{}} & \textbf{71.84\std{1.94}} & 90.99\std{0.33} \\
\bottomrule
\end{tabular}
\subcaption{Discovery baselines, BERT DFR (\%).}
\label{tab:discovery_baselines}
\end{subtable}
\caption{CivilComments-WILDS 16-cell worst-group accuracy on the WILDS test split, 3-seed
mean\std{std} over seeds $\{42,123,323\}$. \textbf{(a)} Debiasing methods. BERT uses Tier~A
(9 causally verified features), RoBERTa uses Tier~C (all 10 SFV features). JTT is excluded
because the BERT ERM error rate (${\approx}3\%$) is too low for upweighting to
converge~\citep{liu2021justtraintwiceimproving}. Kirichenko~'23 uses hand-labeled groups.
\textbf{(b)} Discovery baselines under a fixed BERT DFR recipe, so that only feature
discovery differs.}
\label{tab:cc_combined}
\end{table}

\paragraph{Comparison against simpler discovery.}
If cheaper feature mining produced the same groups, the discovery stage would not be
earning the gains above. We therefore compare three discovery baselines while holding the
downstream debiasing recipe fixed, so that discovery is the only difference:
\textbf{PMI} (top-10 single-token statistics, \citealp{gururangan2018annotationartifactsnaturallanguage}),
\textbf{PMI+SFV} (PMI candidates passed through our statistical validator), and
\textbf{LLM-only} (prompt-only LLM candidates with no validation funnel).
Table~\ref{tab:discovery_baselines} reports the comparison on CivilComments-WILDS.

\modelname{} exceeds the strongest baseline (PMI+SFV) by 3.65\,pp WGA with non-overlapping
mean\,$\pm$\,std intervals, while staying within 2\,pp of every baseline on average
accuracy. Because PMI+SFV passes through the \emph{same} statistical validator, the residual
gap isolates the value of compositional LLM-generated candidates over single-token
statistics. The gap is visible without training anything: PMI's top-10 is entirely insult
vocabulary and recovers \emph{none} of the eight canonical WILDS identity axes, while
\modelname{} recovers six of eight with no access to demographic labels.

The ordering is not uniform across debiasing methods. Under PoE, \modelname{} reaches
65.96\std{0.99} WGA against PMI's 69.34\std{1.09}, trailing the lexical baselines by
3.38\,pp. The 16-cell decomposition in Appendix~\ref{app:cc_poe_cells} shows the same sign on every
axis: \modelname{}'s PoE is more accurate on all eight non-toxic identity cells and less
accurate on all eight toxic ones. We read this as heavier suppression of the
identity--toxicity correlation, which lowers false positives on identity-mentioning speech
and raises false negatives on toxic content aimed at those identities. The corresponding NLI
comparison is in Appendix~\ref{app:nli_baselines}.

\subsection{Ablation Studies}
\label{sec:ablations}
\label{sec:sentiment_ablation}

\paragraph{Sentiment: causal verification as a gate.}
Sentiment supplies a controlled setting in which verification should, if it is working,
return nothing. On SST-2 and IMDB with BERT-base-uncased (seed 42), SFV identifies ten
high-confidence lexical features per corpus by conventional criteria, with mean odds
ratios of 4.94 on SST-2 and 5.14 on IMDB and mean precision above 0.80 on both. Of these,
5 on SST-2 and 8 on IMDB carry enough feature-present coverage to test causally, and
verification returns \textbf{0 of 5 and 0 of 8 exploited respectively} (mean
$\bar{\Delta p} \approx -0.0026$ and $-0.0067$, with no paired $t$-test significant at
$\alpha = 0.05$). We expect this outcome. Sentiment words are the surface
realization of the gold label, so they correlate with the class because they \emph{are}
what the task measures. Lexical overlap and negation in NLI are different: there the
surface feature is logically independent of the inference relation. Consistent with that,
debiasing IMDB on these statistically-validated-but-non-causal features leaves
in-distribution accuracy essentially unchanged (ERM 92.07 against DFR 91.85, PoE 92.04).
Full statistics are in Appendix~\ref{app:sentiment_ablation_full} (single seed, single architecture).

\paragraph{Causal verification vs.\ statistical filtering (Tier~A vs.\ Tier~C).}
On CivilComments, we compare debiasing with all 10 SFV-validated features (Tier~C) against the 9 that also pass single-target verification (Tier~A), which excludes the one feature not individually sufficient on BERT.
For BERT, Tier~A improves DFR worst-group accuracy by $+$1.17\,pp over Tier~C (71.84 vs.\ 70.67): for the DFR family a single non-exploited feature in the group partition adds noise rather than signal. For PoE, SCER and LEACE the two tiers differ by under 0.5\,pp.
RoBERTa shows no consistent Tier~A benefit. With stronger representations the full 10-feature set provides equally informative group partitions.
Full Tier~A and Tier~C results across all methods are in Appendix~\ref{app:cc_full}, and full pipeline funnel statistics for both tasks are in Appendix~\ref{app:pipeline_stats}.

\paragraph{Human evaluation of counterfactual quality.}
We select the \textsc{EvaluatorLLM} using ChaosNLI, which provides multiple human annotations per example rather than a single gold label. On 1,350 samples Qwen3-32B achieves 80.3\% agreement with the human-majority vote, and the six-model comparison is in Appendix~\ref{app:model_selection}. Because ChaosNLI is built from high-disagreement items, that figure is a worst-case calibration bound rather than a measure of the counterfactuals the causal stage actually consumes. A direct audit of 100 accepted counterfactuals gives 94/100 gold-label preservation (Appendix~\ref{app:cf_audit}).

\subsection{Qualitative Analysis on RewardBench2}
\label{sec:rb2_qualitative}

To demonstrate generalizability beyond NLI, we apply the discovery and validation stages (\S\ref{sec:scdata}) to RewardBench2~\citep{malik2025rewardbench2advancingreward}, a human preference benchmark covering six task subsets. Discovery and validation run here without a target model to verify against. We run the full discovery pipeline independently on each subset, treating each (prompt, chosen, rejected) triple as a binary preference instance under the same SFV configuration as NLI. Table~\ref{tab:rb2_features} summarizes the findings.

The pipeline recovers known reward-model biases without being told to look for them.
\textbf{Focus} is dominated by length and formatting on the rejected side, where nine of
ten top features are non-lexical, independently quantifying the format bias documented
by~\citet{zhang2025listsemojisformatbias}. \textbf{Safety} recovers a refusal-language
hierarchy peaking at OR\,=\,161 (precision\,=\,1.00) for apologetic refusals, matching the
sycophancy and rule-based reward patterns of~\citet{sharma2025understandingsycophancylanguagemodels}
and~\citet{mu2024rulebasedrewardslanguage}. \textbf{Ties}, despite its label-neutral
construction, remains length-sensitive (OR\,=\,2.0), recovering the length--reward
correlation of~\citet{singhal2024longwaygoinvestigating}. \textbf{Math} rewards a didactic
opener over correctness, and \textbf{Factuality} turns on markdown headers.

Most of these features are structural -- length thresholds, list-item counts, and counts of
formatting markers -- which n-gram enumeration cannot express, the same coverage gap that
separates \modelname{} from single-token discovery in Table~\ref{tab:discovery_baselines}.
\textbf{Precise\_IF} is the one subset where nothing survives, with zero significant features
across 95 candidates, and discovery alone cannot tell whether it carries no surface signal or
signal our predicates cannot express. Representative features per subset are in Appendix Table~\ref{tab:rb2_features},
and full lists in Appendix~\ref{app:rb2_full}.

\section{Conclusion}
\modelname{} bridges dataset-level statistical artifacts and model-level shortcut
exploitation through three auditable stages: automated discovery, causal verification, and
annotation-free mitigation. The boolean-expression representation is the key design choice,
since one deterministic function serves as discovery target, counterfactual intervention
handle, and group-label generator. Verification surfaces a cross-architecture divergence
invisible to correlation-only analysis, and on sentiment it correctly returns zero exploited
features, confirming its role as a decision gate rather than a refinement. On CivilComments,
programmatic groups match hand-labeled DFR without demographic annotation, and the discovery
stages generalize to RewardBench2 preference data unmodified.

\section{Limitations}
Requiring every feature to be an executable boolean expression bounds what \modelname{} can
discover. Latent and semantic shortcuts, topic and style bias, and distributional artifacts
such as class imbalance cannot be written as a deterministic predicate over the input, so
the pipeline does not see them. The causal stage is bounded by its generator, since a
counterfactual edit removes the target pattern but cannot hold every other property of the
input fixed, and because co-firing features are removed together, $\Delta p$ measures the
effect of an edit and upper-bounds each feature's individual contribution.

\section*{Reproducibility Statement}
All hyperparameters, prompt templates, and pipeline configuration details are in
Appendix~\ref{app:hyperparams}. Source code for the discovery, causal verification, and
debiasing stages, together with the per-sample counterfactual annotation sheets underlying
the audit in Appendix~\ref{app:cf_audit}, is released at
\url{https://github.com/chidaksh/spurious_mitigator}. All reported NLI and
CivilComments results use three fixed seeds (42, 123, 323). The SFV feature set is held
constant across seeds so that only model-dependent stages vary. Appendix~\ref{app:open_weight} reports a full re-run of the discovery pipeline with an
open-weight generator, for settings where the proprietary endpoint is unavailable. Feature
evaluation is fully programmatic and reproducible without API access. Regenerating
counterfactuals requires
GPT-4o and Qwen3-32B, at the cost reported in Appendix~\ref{app:hyperparams}.

\bibliography{references}
\bibliographystyle{colm2026_conference}

\appendix
\raggedbottom
\setlength{\intextsep}{8pt plus 2pt minus 2pt}
\setlength{\textfloatsep}{8pt plus 2pt minus 2pt}
\section*{Appendix}
\section{Pipeline Details and Design Choices}
\label{app:pipeline}

\subsection{NLI Spurious Feature Discovery}
\label{app:spurious_features}

Table~\ref{tab:spurious_features} lists the 10 SFV-selected features on MNLI alongside their statistical and causal properties. The three lexical-overlap variants fire on 26--89\% of training examples and carry the largest causal effects ($\bar{\Delta p}$ of $-$11 to $-$15\,pp), confirming heavy entailment-class reliance. Negation-in-hypothesis and contradictory absolute terms anchor contradiction predictions. The hypothesis-length bias is the only structural pattern, associated with the neutral class. Six features are causally verified on both architectures. Three additional contradiction-class features are exploited by BERT only.

\begin{table}[H]
\centering
\footnotesize
\setlength{\tabcolsep}{3pt}
\begin{tabular}{p{5.5cm}lcccccc}
\toprule
 & & & & & \multicolumn{2}{c}{\textbf{Causal}} \\
\cmidrule(lr){6-7}
\textbf{Pattern} & \textbf{Label} & \textbf{OR} & \textbf{Cov.} & $\bar{\boldsymbol{\Delta p}}$ & \textbf{B} & \textbf{R} \\
\midrule
\multicolumn{7}{l}{\textbf{\textit{Causally exploited by both architectures in all seeds (3/3)}}} \\[2pt]
Premise-hyp.\ word overlap ${>}$2         & Ent. & 3.49 & \textbf{88.5\%} & \textbf{$-$15.0\std{3.2}} & \ding{51}\ding{51}\ding{51} & \ding{51}\ding{51}\ding{51} \\
Word overlap (${>}$50\%)             & Ent. & 3.18 & 38.7\% & $-$13.2\std{3.0} & \ding{51}\ding{51}\ding{51} & \ding{51}\ding{51}\ding{51} \\
High token overlap ratio (${>}$40\%)  & Ent. & 3.08 & 26.2\% & $-$11.4\std{1.0} & \ding{51}\ding{51}\ding{51} & \ding{51}\ding{51}\ding{51} \\
Negation words in hypothesis                & Con. & 2.86 & 31.6\% & $-$7.9\std{1.7}  & \ding{51}\ding{51}\ding{51} & \ding{51}\ding{51}\ding{51} \\
Contradictory absolutes in prem.\ \& hyp. & Con. & 5.58 & 6.4\%  & $-$8.6\std{2.7}  & \ding{51}\ding{51}\ding{51} & \ding{51}\ding{51}\ding{51} \\
Hypothesis ${>}$1.5$\times$ premise length    & Neu. & 3.26 & 9.1\%  & $-$11.1\std{2.8} & \ding{51}\ding{51}\ding{51} & \ding{51}\ding{51}\ding{51} \\
\midrule
\multicolumn{7}{l}{\textbf{\textit{BERT only (RoBERTa immune across all seeds)}}} \\[2pt]
Existential (P) + negation (H)              & Con. & 5.39 & 4.2\%  & $-$6.8\std{1.3}  & \ding{51}\ding{51}\ding{51} & \ding{55}\ding{55}\ding{55} \\
Conjunction (P) + absolute (H)              & Con. & 3.35 & 2.0\%  & $-$6.7\std{1.1}  & \ding{51}\ding{51}\ding{51} & \ding{55}\ding{55}\ding{55} \\
Opposite meaning absolutes in prem.\ \& hyp.  & Con. & 8.05 & 1.6\%  & $-$10.1\std{1.4} & \ding{51}\ding{51}\ding{51} & \ding{55}\ding{55}\ding{55} \\
\midrule
\multicolumn{7}{l}{\textbf{\textit{Not causally exploited by either architecture (0/3)}}} \\[2pt]
always/every (P) \& never/no (H) & Con. & \textbf{10.01} & 1.0\% & $-$7.4\std{1.8} & \ding{55}\ding{55}\ding{55} & \ding{55}\ding{55}\ding{55} \\
\bottomrule
\end{tabular}
\caption{Spurious features identified by the pipeline on MNLI. OR = odds ratio on the SFV validation split (all BH-corrected $p < 0.05$). Cov. = fraction of samples with the associated label where the boolean pattern fires. $\bar{\Delta p}$ = mean $\pm$ std change in class probability under removal counterfactuals across 3 seeds (\%). \textbf{B}/\textbf{R} = per-seed causal verdict for BERT-base-uncased / RoBERTa-base. Ent./Con./Neu.\ = Entailment/Contradiction/Neutral.}
\label{tab:spurious_features}
\end{table}

\subsection{Pipeline Funnel Statistics}
\label{app:pipeline_stats}

Table~\ref{tab:pipeline_funnel} reports feature counts at each stage of the pipeline, distinguishing model-independent shared stages from per-seed model-dependent stages. On CivilComments the two deduplication passes remove redundancy at different points: the description-only pass cuts 1{,}695 raw candidates to 506, and the coverage-aware pass applied after boolean logic validation removes a further 47\% (418 to 223) once realized coverage is available to compare.

\begin{table}[H]
\centering
\small
\begin{tabular}{lcc}
\toprule
\textbf{Stage} & \textbf{NLI} & \textbf{CC-WILDS} \\
\midrule
\multicolumn{3}{l}{\textit{Shared (model-independent)}} \\[2pt]
Candidate patterns (\textsc{SCGenLLM})              & 625 & 1{,}695 \\
Post initial deduplication ($\mathcal{SC}_{\mathrm{init}}$)  & 395 & 506 \\
Post boolean logic validation ($\mathcal{SC}_{\mathrm{valid}}$) & 301 & 418 \\
Post coverage-aware deduplication ($\mathcal{SC}_{\mathrm{dist}}$) & 172 & 223 \\
SFV top-$k$ selected ($\mathcal{F}_{\mathrm{val}}$) & 10 & 10 \\
\midrule
\multicolumn{3}{l}{\textit{BERT-base-uncased (model-dependent)}} \\[2pt]
Passed reliance screening ($\mathcal{F}_{\mathrm{suf}}$) & 10 & 10 \\
Causally verified - Tier C ($\mathcal{F}_{\mathrm{causal}}$) & 9 & 10 \\
Causally verified - Tier A (subset) & - & 9 \\
\midrule
\multicolumn{3}{l}{\textit{RoBERTa-base (model-dependent)}} \\[2pt]
Passed reliance screening ($\mathcal{F}_{\mathrm{suf}}$) & 10 & 10 \\
Causally verified - Tier C ($\mathcal{F}_{\mathrm{causal}}$) & 6 & 10 \\
\bottomrule
\end{tabular}
\caption{Feature counts at each pipeline stage for NLI (MNLI) and CivilComments-WILDS (CC-WILDS). Shared stages run once per task, model-dependent stages run per architecture and seed. NLI -- F150 fails causal verification on both architectures across all seeds, F91, F99, F130 additionally fail on RoBERTa (6 verified). CC-WILDS -- all 10 SFV features are causally exploited under joint removal (Tier C), and the 9-feature Tier A subset excludes the one feature that does not additionally pass single-target verification on BERT. Demographic identity coverage -- 6 of the 10 CC-WILDS features contain WILDS identity tokens, together recovering 6 of the 8 canonical identities without any demographic labels.}
\label{tab:pipeline_funnel}
\end{table}

\subsection{Feature Statistics Across Pipeline Stages}
\label{app:feature_stats}

Table~\ref{tab:sfv_stats} reports the SFV validation statistics for all 10 top-$k$ features. Features are listed in rank order (by score). All odds ratios are BH-corrected significant ($p < 0.05$). Table~\ref{tab:boolean_exprs} lists the full executable boolean expressions for each feature.

\begin{table}[H]
\centering
\small
\setlength{\tabcolsep}{4pt}
\resizebox{\textwidth}{!}{%
\begin{tabular}{clccccccc}
\toprule
\textbf{Idx} & \textbf{Pattern (short)} & \textbf{Label} & \textbf{OR} & \textbf{Prec.} & \textbf{Cov.} & \textbf{BH-$p$} & \textbf{Score} & \textbf{Cluster} \\
\midrule
150 & always/every (P) and never/no (H) & C & 10.01 & 0.83 & 0.010 & 0.018 & 0.615 & 2 \\
2   & Premise-hypothesis word overlap                                   & E & 3.49  & 0.39 & 0.885 & $<$0.001 & 0.573 & 2 \\
130 & Opposite-meaning absolutes in both fields                          & C & 8.05  & 0.80 & 0.016 & 0.003 & 0.567 & 1 \\
160 & Contradictory absolute terms cross-sentence                        & C & 5.58  & 0.73 & 0.064 & $<$0.001 & 0.494 & 2 \\
91  & Existential (P) + negation (H)                                     & C & 5.39  & 0.72 & 0.042 & $<$0.001 & 0.479 & 1 \\
16  & Majority word overlap ($>$50\%)                                    & E & 3.18  & 0.54 & 0.387 & $<$0.001 & 0.440 & 1 \\
72  & High token overlap ratio ($>$0.4 Jaccard)                         & E & 3.08  & 0.56 & 0.262 & $<$0.001 & 0.399 & 2 \\
25  & Negation words in hypothesis                                       & C & 2.86  & 0.53 & 0.316 & $<$0.001 & 0.394 & 2 \\
52  & Hypothesis $>$1.5$\times$ premise length                          & N & 3.26  & 0.60 & 0.091 & $<$0.001 & 0.365 & 1 \\
99  & Conjunction (P) + absolute (H)                                     & C & 3.35  & 0.62 & 0.020 & 0.029 & 0.355 & 1 \\
\bottomrule
\end{tabular}}
\caption{SFV validation statistics for the 10 top-$k$ selected features. \textbf{OR} = odds ratio on the validation split (all BH-corrected $p<0.05$). \textbf{Prec.} = $P(y{=}y_k \mid b_i{=}1)$. \textbf{Cov.} = $P(b_i{=}1 \mid y{=}y_k)$. \textbf{Score} = composite of $\tfrac{1}{3}\log\mathrm{OR} + \tfrac{1}{3}\mathrm{Cov} + \tfrac{1}{3}\mathrm{Prec}$ (log-OR min-max normalized). \textbf{Cluster} = number of patterns in the co-occurrence cluster from which this feature was selected as representative. Label = E = Entailment, C = Contradiction, N = Neutral.}
\label{tab:sfv_stats}
\end{table}

\begin{table}[H]
\centering
\footnotesize
\setlength{\tabcolsep}{3pt}
\begin{tabular}{clL{11.6cm}}
\toprule
\textbf{Idx} & \textbf{Label} & \textbf{Boolean Expression} \\
\midrule
150 & C & \texttt{any(w in premise.split() for w in ['always', 'all', 'every']) and any(w in hypothesis.split() for w in ['never', 'none', 'no'])} \\[4pt]
2   & E & \texttt{len(set(re.sub(r'\textbackslash W+', '\ ', premise.lower()).split()).intersection( set(re.sub(r'\textbackslash W+', '\ ', hypothesis.lower()).split()))) > 2} \\[4pt]
130 & C & \texttt{any(w in premise.lower().split() for w in ['all', 'every', 'always', 'no', 'none', 'never', 'nothing']) and any(w in hypothesis.lower().split() for w in ['all', 'every', 'always', 'no', 'none', 'never', 'nothing']) and (any(w in premise.lower().split() for w in ['no', 'none', 'never', 'nothing']) != any(w in hypothesis.lower().split() for w in ['no', 'none', 'never', 'nothing']))} \\[4pt]
160 & C & \texttt{any(w in premise.split() for w in ['never', 'none', 'nothing', 'nowhere']) != any(w in hypothesis.split() for w in ['never', 'none', 'nothing', 'nowhere'])} \\[4pt]
91  & C & \texttt{any(w in premise.split() for w in ['there', 'was', 'is', 'are']) and any(w in hypothesis.split() for w in ['no', 'none'])} \\[4pt]
16  & E & \texttt{len(set(premise.split()).intersection(hypothesis.split())) / min(len(premise.split()), len(hypothesis.split())) > 0.5} \\[4pt]
72  & E & \texttt{len(set(re.findall(r'\textbackslash w+', premise.lower())).intersection( set(re.findall(r'\textbackslash w+', hypothesis.lower())))) / len(set(re.findall(r'\textbackslash w+', premise.lower())).union( set(re.findall(r'\textbackslash w+', hypothesis.lower())))) > 0.4} \\[4pt]
25  & C & \texttt{any(w in hypothesis.split() for w in ['never', 'no', 'not', 'none', 'nothing', "can't", "don't", "won't", "isn't", "aren't", "wasn't", "weren't", "hasn't", "haven't", "hadn't", "doesn't", "didn't", "couldn't", "shouldn't", "wouldn't", "mustn't"])} \\[4pt]
52  & N & \texttt{len(re.findall(r'\textbackslash w+', hypothesis)) > len(re.findall(r'\textbackslash w+', premise)) * 1.5 if len(premise) > 0 else False} \\[4pt]
99  & C & \texttt{any(w in premise.split() for w in ['and', 'but', 'however']) and any(w.strip('.,!?') in hypothesis.split() for w in ['never', 'always'])} \\
\bottomrule
\end{tabular}
\caption{Executable boolean expressions for all 10 top-$k$ features. Each expression evaluates deterministically on any NLI pair (\texttt{premise}, \texttt{hypothesis}) and serves as the evaluation backbone for all downstream stages. Backslash characters are escaped for display. Expressions use Python syntax with \texttt{re} imported.}
\label{tab:boolean_exprs}
\end{table}

\clearpage

\subsection{CivilComments Spurious Feature Discovery}
\label{app:cc_features}

Table~\ref{tab:cc_spurious_features} lists the 10 SFV-selected features on CivilComments-WILDS alongside their causal verification results. Six features contain tokens from WILDS's 8 canonical demographic identity attributes and a seventh is thematically related without containing one. Three are novel patterns absent from the annotation schema. All 10 are causally exploited by both architectures across all seeds (Tier~C). On BERT, single-target verification excludes F120 from Tier~A (the interjection feature's surface cue is not individually sufficient to shift predictions, though it contributes in combination with other features).

\begin{table}[H]
\centering
\footnotesize
\setlength{\tabcolsep}{3pt}
\begin{tabular}{clL{4.2cm}rr}
\toprule
 & & & \multicolumn{2}{c}{\textbf{Causal $\bar{\boldsymbol{\Delta p}}$}} \\
\cmidrule(lr){4-5}
\textbf{Idx} & \textbf{Pattern} & \textbf{WILDS Identity Overlap} & \textbf{BERT} & \textbf{RoBERTa} \\
\midrule
\multicolumn{5}{l}{\textbf{\textit{Overlapping WILDS demographic identities (7 features)}}} \\[2pt]
24  & Religious keywords                         & christian, muslim, other\_religions & $-$43.0 & $-$42.6 \\
40  & Racial or ethnic identifiers               & black, white                        & $-$48.5 & $-$47.8 \\
53  & Ethnic, religious, or racial group names    & black, white, christian, muslim      & $-$52.1 & $-$51.0 \\
78  & Race or ethnicity mentions                  & black, white                        & $-$46.3 & $-$45.2 \\
82  & Social issues (racism, sexism)              & (gender, thematic)$^{\S}$          & $-$48.3 & $-$48.4 \\
96  & Racism- or supremacy-related terms          & black, white                        & $-$49.5 & $-$49.2 \\
117 & Sexual orientation terms                    & LGBTQ                               & $-$49.4 & $-$47.6 \\
\midrule
\multicolumn{5}{l}{\textbf{\textit{Novel patterns (absent from WILDS annotations)}}} \\[2pt]
59  & Frequent use of ``you''                     & --                                  & $-$39.7 & $-$40.1 \\
120 & Interjections or exclamatory words$^{\dagger}$ & --                              & $-$24.3 & $-$28.4 \\
127 & Derogatory terms (``idiot'', ``stupid'')    & --                                  & \textbf{$-$66.8} & \textbf{$-$65.6} \\
\bottomrule
\end{tabular}
\caption{Spurious features identified by the pipeline on CivilComments-WILDS. All 10 features are causally exploited by both architectures across all 3 seeds under joint removal ($p < 10^{-4}$ for all). $\bar{\Delta p}$ = mean percentage-point change in toxic-class probability under removal counterfactuals (BERT seed~123 shown, other seeds are consistent). WILDS overlap determined post-hoc by token matching against the 8 canonical identity attributes. Token matching recovers 6 of the 8, with \emph{male} and \emph{female} unrecovered. $^{\S}$F82 is thematically about gender but contains no identity token, so it is not counted among the 6. F127 (derogatory terms) has the largest causal effect on both architectures. $^{\dagger}$F120 is excluded from Tier~A on BERT (single-target $p = 0.068$).}
\label{tab:cc_spurious_features}
\end{table}

\subsection{Sentiment Feature Discovery (IMDB and SST-2)}
\label{app:sentiment_features}

Tables~\ref{tab:imdb_sfv_stats} and~\ref{tab:sst2_sfv_stats} report the 10 SFV-selected features for IMDB and SST-2 respectively. All features are sentiment-lexicon patterns (positive and negative adjectives, superlatives, critique vocabulary) that pass statistical validation with $\text{OR} \geq 4.3$ on IMDB and $\geq 2.7$ on SST-2. Causal verification returns zero exploited features on both datasets: the mean $|\Delta p|$ across all tested features is 0.67\,pp on IMDB ($n{=}8$ tested, 0/8 exploited) and 0.26\,pp on SST-2 ($n{=}5$ tested, 0/5 exploited). This confirms the pipeline's causal gate correctly filters dataset-level lexical correlations that models have learned to ignore. Tables~\ref{tab:imdb_boolean_exprs} and~\ref{tab:sst2_boolean_exprs} list the executable boolean expressions.

\begin{table}[H]
\centering
\small
\setlength{\tabcolsep}{4pt}
\resizebox{\textwidth}{!}{%
\begin{tabular}{clccccccc}
\toprule
\textbf{Idx} & \textbf{Pattern (short)} & \textbf{Label} & \textbf{OR} & \textbf{Prec.} & \textbf{Cov.} & \textbf{BH-$p$} & \textbf{Score} & \textbf{Cluster} \\
\midrule
55  & Strongly negative adjectives (``stupid'', ``crap'')    & Neg. & 6.67 & 0.84 & 0.230 & $<$0.001 & 0.691 & 1 \\
27  & Common negative adjectives (``bad'', ``worst'')        & Neg. & 4.88 & 0.75 & 0.487 & $<$0.001 & 0.676 & 2 \\
62  & Specific negative words (``disappoint'', ``bored'')    & Neg. & 5.18 & 0.81 & 0.212 & $<$0.001 & 0.619 & 1 \\
31  & Overly positive adjectives (``awesome'', ``brilliant'') & Pos. & 4.28 & 0.75 & 0.396 & $<$0.001 & 0.615 & 1 \\
63  & Strong positive adjectives (``superb'', ``excellent'') & Pos. & 5.26 & 0.82 & 0.179 & $<$0.001 & 0.612 & 1 \\
88  & Negative critique words (``laughable'', ``stupid'')    & Neg. & 5.54 & 0.84 & 0.096 & $<$0.001 & 0.603 & 1 \\
101 & ``favorite/favorites'' with negation context           & Pos. & 5.67 & 0.84 & 0.064 & $<$0.001 & 0.599 & 1 \\
97  & ``amazing'', ``astonishing'', ``astonished''           & Pos. & 5.25 & 0.83 & 0.072 & $<$0.001 & 0.581 & 1 \\
66  & Hyperbolic adjectives (``brilliant'', ``incredible'')  & Pos. & 4.32 & 0.79 & 0.174 & $<$0.001 & 0.557 & 1 \\
68  & Specific positive descriptors (``excellent'', ``fantastic'') & Pos. & 4.36 & 0.79 & 0.158 & $<$0.001 & 0.555 & 1 \\
\bottomrule
\end{tabular}}
\caption{SFV validation statistics for the 10 top-$k$ features on IMDB. Column definitions match Table~\ref{tab:sfv_stats}. All features are sentiment-lexicon patterns. Causal verification returns 0/8 exploited (mean $|\Delta p| = 0.67$\,pp), with 2 features excluded from testing due to coverage constraints.}
\label{tab:imdb_sfv_stats}
\end{table}

\begin{table}[H]
\centering
\small
\setlength{\tabcolsep}{4pt}
\resizebox{\textwidth}{!}{%
\begin{tabular}{clccccccc}
\toprule
\textbf{Idx} & \textbf{Pattern (short)} & \textbf{Label} & \textbf{OR} & \textbf{Prec.} & \textbf{Cov.} & \textbf{BH-$p$} & \textbf{Score} & \textbf{Cluster} \\
\midrule
116 & Comparison/insufficiency (``not enough'', ``too much'') & Neg. & 14.51 & 1.00 & 0.008 & 0.015 & 0.670 & 1 \\
62  & Superlative language (``best'', ``terrific'', ``perfect'') & Pos. & 4.93 & 0.86 & 0.022 & 0.004 & 0.452 & 1 \\
79  & ``entertaining'', ``outstanding'', ``enjoyable''       & Pos. & 4.13 & 0.84 & 0.019 & 0.018 & 0.415 & 1 \\
102 & Negative adjectives (``worst'', ``cheap'', ``lazy'')   & Neg. & 4.62 & 0.78 & 0.011 & 0.045 & 0.409 & 1 \\
120 & Strong negative adjectives (``horrible'', ``awful'')   & Neg. & 4.62 & 0.78 & 0.011 & 0.045 & 0.409 & 1 \\
65  & Positive adjectives (``great'', ``excellent'', ``admirable'') & Pos. & 3.90 & 0.83 & 0.029 & 0.003 & 0.406 & 1 \\
74  & ``love'', ``enjoy'', ``liked''                         & Pos. & 3.87 & 0.83 & 0.018 & 0.029 & 0.401 & 1 \\
59  & Positive-sounding words (``great'', ``lovely'')        & Pos. & 3.25 & 0.81 & 0.029 & 0.006 & 0.367 & 1 \\
58  & Positive adjectives (``unique'', ``spectacular'')      & Pos. & 2.89 & 0.79 & 0.031 & 0.012 & 0.342 & 1 \\
56  & Positive adjectives (``best'', ``brilliant'', ``engaging'') & Pos. & 2.72 & 0.78 & 0.025 & 0.030 & 0.326 & 1 \\
\bottomrule
\end{tabular}}
\caption{SFV validation statistics for the 10 top-$k$ features on SST-2. Column definitions match Table~\ref{tab:sfv_stats}. F116 has the highest OR (14.51) but the lowest coverage (0.8\%), consistent with a rare but strong lexical pattern. F102 and F120 fire on the same validation examples and so share every statistic, and were retained separately because the coverage-aware merge requires description similarity as well as coverage overlap. Causal verification returns 0/5 exploited (mean $|\Delta p| = 0.26$\,pp), with 5 features excluded due to SST-2's shorter sentence lengths reducing feature-present pool sizes.}
\label{tab:sst2_sfv_stats}
\end{table}

\begin{table}[H]
\centering
\footnotesize
\setlength{\tabcolsep}{3pt}
\begin{tabular}{clL{11.6cm}}
\toprule
\textbf{Idx} & \textbf{Label} & \textbf{Boolean Expression} \\
\midrule
55  & Neg. & \texttt{any(re.search(r'\textbackslash b' + re.escape(w) + r'\textbackslash b', text, re.IGNORECASE) for w in ['stupid', 'crap', 'sucks', 'terrible'])} \\[4pt]
27  & Neg. & \texttt{any(re.search(r'\textbackslash b' + w + r'\textbackslash b', text) for w in ['bad', 'worst', 'terrible']) and not re.search(r'\textbackslash bnot\textbackslash s+(?:necessarily\textbackslash s+)?\allowbreak(?:that\textbackslash s+)?(?:bad|worst|terrible)\textbackslash b', text)} \\[4pt]
62  & Neg. & \texttt{any(w in text.split() for w in ['worst', 'disappoint', 'bored', 'terrible'])} \\[4pt]
31  & Pos. & \texttt{any(re.search(r'\textbackslash b' + w + r'\textbackslash b', text, re.IGNORECASE) for w in ['awesome', 'brilliant', 'fantastic', 'incredible', 'amazing', 'excellent', 'superb', 'perfect', 'flawless', 'hilarious'])} \\[4pt]
63  & Pos. & \texttt{any(w in text.split() for w in ['superb', 'excellent', 'wonderful'])} \\[4pt]
88  & Neg. & \texttt{any(w in text.split() for w in ['laughable', 'stupid', 'crap'])} \\[4pt]
101 & Pos. & \texttt{any(w in text.split() for w in ['favorite', 'favorites']) and any(neg in text for neg in ['not', 'no', 'terrible', 'worst', 'bad'])} \\[4pt]
97  & Pos. & \texttt{re.search(r'\textbackslash b(?:amazing|astonishing|astonished)\textbackslash b', text, re.IGNORECASE)} \\[4pt]
66  & Pos. & \texttt{any(phrase in text for phrase in ['brilliant', 'incredible', 'amazing'])} \\[4pt]
68  & Pos. & \texttt{any(word in text.split() for word in ['excellent', 'fantastic', 'amazing'])} \\
\bottomrule
\end{tabular}
\caption{Executable boolean expressions for all 10 top-$k$ IMDB features. Each expression evaluates on a single \texttt{text} field. Despite strong statistical associations (OR $\geq$ 4.3), none of these lexical patterns are causally exploited by BERT on IMDB (mean $|\Delta p| = 0.67$\,pp), confirming that sentiment models encode task-relevant signal rather than surface heuristics.}
\label{tab:imdb_boolean_exprs}
\end{table}

\begin{table}[H]
\centering
\footnotesize
\setlength{\tabcolsep}{3pt}
\begin{tabular}{clL{11.6cm}}
\toprule
\textbf{Idx} & \textbf{Label} & \textbf{Boolean Expression} \\
\midrule
116 & Neg. & \texttt{any(phrase in text for phrase in ['not enough', 'too much', 'little more', 'less than', 'little insight', 'not as'])} \\[4pt]
62  & Pos. & \texttt{any(w in text.split() for w in ['best', 'terrific', 'perfect', 'spectacularly', 'stellar', 'perfection'])} \\[4pt]
79  & Pos. & \texttt{any(w in text.split() for w in ['entertaining', 'outstanding', 'enjoyable'])} \\[4pt]
102 & Neg. & \texttt{any(w in text.split() for w in ['worst', 'cheap', 'lazy'])} \\[4pt]
120 & Neg. & \texttt{any(w in text.split() for w in ['horrifying', 'awfulness', 'awful', 'horrible', 'terrible', 'offensive'])} \\[4pt]
65  & Pos. & \texttt{any(w in text.split() for w in ['great', 'excellent', 'fine', 'admirable', 'promisingly'])} \\[4pt]
74  & Pos. & \texttt{any(w in text.split() for w in ['love', 'enjoy', 'liked'])} \\[4pt]
59  & Pos. & \texttt{any(w in text.split() for w in ['great', 'lovely', 'interesting'])} \\[4pt]
58  & Pos. & \texttt{any(w in text.split() for w in ['unique', 'spectacular', 'great', 'fascinating'])} \\[4pt]
56  & Pos. & \texttt{any(w in text.split() for w in ['best', 'lovely', 'engaging', 'exciting', 'brilliant'])} \\
\bottomrule
\end{tabular}
\caption{Executable boolean expressions for all 10 top-$k$ SST-2 features. SST-2's shorter sentences (mean 19 tokens vs.\ IMDB's 231) concentrate features around individual sentiment words rather than compound expressions.}
\label{tab:sst2_boolean_exprs}
\end{table}

\subsection{Sentiment Ablation: Full Debiasing Grid}
\label{app:sentiment_ablation_full}

Table~\ref{tab:sentiment_debias} reports what happens when causal verification is bypassed
and IMDB is debiased directly on the eight features that passed sufficiency screening but
that verification found to be non-causal (\S\ref{sec:sentiment_ablation}). Seven of the
eight methods stay within $\pm$0.5\,pp of the ERM baseline in distribution, while
out-of-distribution accuracy is flat or degraded. LEACE is the exception in the opposite
direction: erasing the full sentiment subspace destroys the distributed representation the
task depends on, which is a coarser intervention than the token-level removal that
$\Delta p$ measures. SCER and DFR each improve marginally on one out-of-distribution set
(Yelp and TweetEval, and SST-2 respectively), so the pattern is a null-to-negative
result rather than a uniform degradation.
These runs use a single seed and a single architecture.

\begin{table}[H]
\centering
\small
\setlength{\tabcolsep}{4pt}
\begin{tabular}{lcccc}
\toprule
\textbf{Method} & \textbf{IMDB} & \textbf{SST-2} & \textbf{Yelp} & \textbf{TweetEval} \\
\midrule
ERM    & 92.07 & 87.84 & 91.41 & 78.95 \\
DFR    & 91.85 & \textbf{88.99} & 90.66 & 75.28 \\
SCER   & 92.02 & 87.61 & \textbf{91.77} & \textbf{80.10} \\
LEACE  & 72.22 & 84.17 & 79.94 & 62.75 \\
\bottomrule
\end{tabular}
\caption{IMDB debiasing accuracy (\%) when applied to 8 features that passed sufficiency screening but have \textbf{0 causally verified features}. IMDB = in-distribution, SST-2/Yelp/TweetEval = OOD. BERT-base-uncased, seed 42. Full 8-method results confirm the same pattern. Selected methods shown for clarity. \textbf{Bold} = marginal OOD gain over ERM.}
\label{tab:sentiment_debias}
\end{table}

\subsection{Counterfactual Sample Sizes and Test Statistics}
\label{app:cf_stats}

Table~\ref{tab:cf_sample_sizes} reports the per-feature counterfactual sample sizes, acceptance rates, and paired $t$-test statistics for the causal verification stage on NLI (MNLI). All 10 SFV features enter the counterfactual generation pipeline. The counterfactual pool per feature is determined by the number of feature-present examples drawn from the associated-label partition of the 5{,}000-sample training subset. Generated counterfactuals are accepted only if the Evaluator~LLM confirms label preservation. The acceptance rate reflects the fraction of pooled examples that yield valid counterfactual pairs. The same accepted counterfactual set is evaluated on both architectures across all three seeds. BERT seed~42 and RoBERTa seed~323 are shown as representative (statistics are stable across seeds for verified features). Feature~F150 is the sole non-exploited feature on both architectures: with $n{=}27$ accepted pairs, its $t$-test is underpowered (post-hoc power $= 0.34$ at $d{=}0.316$), though the small effect size and CI crossing zero are consistent with genuine non-exploitation rather than a Type~II error.

\begin{table}[H]
\centering
\footnotesize
\setlength{\tabcolsep}{2pt}
\begin{tabular}{lcrrrrrrrr}
\toprule
 & & & & & \multicolumn{2}{c}{\textbf{BERT (seed 42)}} & \multicolumn{2}{c}{\textbf{RoBERTa (seed 323)}} \\
\cmidrule(lr){6-7} \cmidrule(lr){8-9}
\textbf{Feature} & \textbf{Label} & $N_{\mathrm{rm}}$ & $N_{\mathrm{cf}}$ & \textbf{Ded.} & $t$ & $p$ & $t$ & $p$ \\
\midrule
\multicolumn{9}{l}{\textbf{\textit{Causally exploited by both architectures}}} \\[2pt]
F2 \; (word overlap ${>}$2)          & Ent. & 107 & 106 & 99.1\% & $-$4.84 & $4.5{\times}10^{-6}$ & $-$4.84 & $4.5{\times}10^{-6}$ \\
F16 (overlap ${>}$50\%)              & Ent. & 197 & 188 & 95.4\% & $-$4.52 & $1.1{\times}10^{-5}$ & $-$4.20 & $4.1{\times}10^{-5}$ \\
F72 (token overlap ${>}$40\%)        & Ent. & 151 & 141 & 93.4\% & $-$3.85 & $1.8{\times}10^{-4}$ & $-$3.53 & $5.6{\times}10^{-4}$ \\
F25 (negation in hyp.)               & Con. & 456 & 421 & 92.3\% & $-$5.75 & $1.7{\times}10^{-8}$ & $-$4.24 & $2.7{\times}10^{-5}$ \\
F160 (contradictory absolutes)       & Con. & 185 & 152 & 82.2\% & $-$4.53 & $1.2{\times}10^{-5}$ & $-$2.71 & $7.5{\times}10^{-3}$ \\
F52 (hyp.\ length ${>}$1.5$\times$)  & Neu. & 93  & 93  & 100\%  & $-$4.16 & $7.3{\times}10^{-5}$ & $-$3.02 & $3.2{\times}10^{-3}$ \\
\midrule
\multicolumn{9}{l}{\textbf{\textit{BERT only (RoBERTa immune)}}} \\[2pt]
F91 (exist.\ + negation)             & Con. & 90  & 78  & 86.7\% & $-$3.13 & $2.5{\times}10^{-3}$ & $-$0.11 & 0.911 \\
F99 (conj.\ + absolute)              & Con. & 104 & 75  & 72.1\% & $-$3.39 & $1.1{\times}10^{-3}$ & $-$1.67 & 0.099 \\
F130 (opposite absolutes)            & Con. & 77  & 47  & 61.0\% & $-$2.81 & $7.3{\times}10^{-3}$ & $-$1.38 & 0.173 \\
\midrule
\multicolumn{9}{l}{\textbf{\textit{Not exploited by either architecture}}} \\[2pt]
F150 (always/every vs.\ never/no)    & Con. & 55  & 27  & 49.1\% & $-$1.64 & 0.113 & $-$1.52 & 0.140 \\
\bottomrule
\end{tabular}
\caption{Per-feature counterfactual sample sizes and paired $t$-test statistics for NLI causal verification. $N_{\mathrm{rm}}$ = accepted counterfactuals in which $b_i$ was programmatically verified removed. $N_{\mathrm{cf}}$ = distinct source examples among them, i.e.\ the paired sample size entering the test once counterfactuals generated from the same original are de-duplicated. Ded.\ = $N_{\mathrm{cf}} / N_{\mathrm{rm}}$, a de-duplication ratio rather than a label-preservation rate. $t$ and $p$ are from the paired $t$-test on $\Delta p$ (change in the model's predicted probability for the spurious label under counterfactual removal). The same counterfactual set is shared across seeds and architectures. Ent./Con./Neu.\ = Entailment/Contradiction/Neutral.}
\label{tab:cf_sample_sizes}
\end{table}

\subsection{Model Design Choices}
\label{app:model_selection}
We evaluate multiple LLMs for two distinct verification roles -- (i)~\textbf{feature presence detection} -- Can an LLM reliably identify whether a spurious feature is present in an NLI sample, verified against a deterministic programmatic check? and (ii)~\textbf{label preservation} -- Can an LLM verify that a generated counterfactual retains the correct NLI label, measured against human annotations?

\paragraph{Feature Presence Detection.}
To probe LLM capability across structurally distinct feature types, we construct 10 diagnostic features spanning four categories -- (A)~\textbf{baseline} patterns (e.g., presence of a keyword or punctuation mark), (B)~\textbf{compound} conditions (e.g., conjunctions or negations of two surface checks), (C)~\textbf{substring} patterns (distinguishing exact-substring from word-boundary matching), and (D)~\textbf{counting} thresholds (e.g., comma count $>$ 2). Each feature has a deterministic programmatic ground truth, enabling exact agreement measurement. Table~\ref{tab:feature_eval} reports agreement by category.

\begin{table}[H]
\centering
\small
\setlength{\tabcolsep}{3pt}
\begin{tabular}{llccc}
\toprule
\textbf{Category} & \textbf{Challenge} & \textbf{Qwen3-32B} & \textbf{Gemini Flash} & \textbf{Flash Lite} \\
\midrule
Baseline (A)  & None               & 99.5 & 97.0 & 97.5 \\
Compound (B)  & AND / negation     & 84.8 & 71.4 & 58.2 \\
Substring (C) & Word boundary      & 96.0 & 91.2 & 87.5 \\
Counting (D)  & Thresholds         & 99.0 & 90.9 & 80.3 \\
\midrule
\textbf{Overall Agreement (\%)} & & \textbf{93.9} & 86.4 & 79.3 \\
\bottomrule
\end{tabular}
\caption{Feature presence agreement (\%) against programmatic ground truth, broken down by feature category ($n{=}99$ samples per feature, 10 features). Compound conditions requiring simultaneous satisfaction of multiple sub-conditions or correct handling of negation are the most challenging category across all models.}
\label{tab:feature_eval}
\end{table}

Although Qwen3-32B performs competitively on LLM-based evaluation, 
we use programmatic boolean expressions throughout as they are 
deterministic, they eliminate API cost for the high-volume SFV stage, 
and avoid the structural failure modes observed in compound and 
counting categories.

\paragraph{Label Preservation.}
We evaluate label preservation accuracy against human-annotated NLI from ChaosNLI (has multiple human annotations unlike other NLI datasets, where mislabelling is a spurious feature in itself) across 1,350 samples (450 per class).

\begin{table}[H]
\centering
\small
\setlength{\tabcolsep}{3pt}
\begin{tabular}{lccc}
\toprule
\textbf{Model} & \textbf{Accuracy} & \textbf{TP} & \textbf{TN} \\
\midrule
Gemini Flash      & \textbf{81.6} & 75.9 & 84.5 \\
Claude            & 80.6          & 73.1 & 84.3 \\
Qwen3-32B         & 80.3          & 72.7 & 84.1 \\
Gemini Pro        & 79.7          & 70.4 & 84.3 \\
Gemini Flash Lite & 79.5          & 73.8 & 82.3 \\
Llama-3.3-70B     & 66.7 & -  & -  \\
\bottomrule
\end{tabular}
\caption{Label preservation accuracy on $n{=}1{,}350$ NLI samples. Accept Correct = rate at which the model accepts a counterfactual whose label matches the gold. Reject Wrong = rate at which the model rejects a mislabelled counterfactual.}
\label{tab:label_eval}
\end{table}

Qwen3-32B is selected as the \textsc{EvaluatorLLM} for both roles. Although Gemini Flash achieves marginally higher label preservation accuracy (81.6\% vs.\ 80.3\%), Qwen3-32B's substantially superior feature presence agreement (93.9\% vs.\ 86.4\%), particularly on compound and counting categories along with the cheaper API costs makes it the preferred unified evaluator. Using a single model for both roles avoids role-specific model management overhead.

For the \textsc{GeneratorLLM}, preliminary experiments showed that GPT-4o-mini produced vague pattern descriptions and frequently generated syntactically malformed or semantically inconsistent boolean expressions. GPT-4o is therefore used for all generation roles (\textsc{SCGenLLM}, \textsc{BLV} rewriter, and CF generator).

\subsection{Human Audit of Accepted Counterfactuals}
\label{app:cf_audit}
The label-preservation accuracy in Table~\ref{tab:label_eval} is measured on
ChaosNLI-MNLI~\citep{nie2020chaosnli}, a benchmark deliberately constructed from
high-human-disagreement examples (100 annotations per item, including near-even label
splits) and evaluated under a stricter protocol of three claimed-label queries per sample.
It is therefore a worst-case calibration figure rather than an estimate of the quality of
the counterfactuals the causal stage actually consumes, which are minimal edits of
single-label MNLI and CivilComments originals.

To measure the latter directly we ran a stratified audit of 100 \textsc{EvaluatorLLM}-accepted
counterfactuals -- 50 NLI and 50 CivilComments, five per SFV-surviving feature -- annotated
by a graduate-level human reviewer. Gold-label preservation was
45/50 (90\%) on NLI and 49/50 (98\%) on CivilComments, i.e.\ \textbf{94/100 (94\%)}
overall, and feature-removal correctness was 98/100 (98\%).\footnote{The NLI figure counts
five borderline cases as failures: instances where the generator's paraphrase narrows or
softens the original (for example \emph{never} to \emph{rarely}) without a gross label
flip. Under a permissive reading that accepts these, NLI label preservation is 50/50. We
report the strict count. Per-sample annotations are released with the code.}
The residual errors are subtle entailment- or contradiction-to-neutral drifts rather than
category changes. The 94\% rate on the accepted-counterfactual subset that the causal stage
consumes therefore substantially exceeds the ChaosNLI worst-case figure.

\subsection{Additional Design Choices}
\label{app:design_choices}

\paragraph{Score formula weighting.}
The SFV score uses equal weights ($0.33$ each) for $\log\mathrm{OR}$, Coverage, and Precision, intentionally preventing statistical significance from dominating selection. Equal weighting still admits very high-OR, low-coverage features: idx 150 (OR$=10.01$, Cov$=1.0\%$) ranks first on score despite firing on under 1\% of training examples, and is precisely the feature causal verification rejects on both architectures. Scoring alone therefore cannot tell whether a model has encoded a rare shortcut, which is what the causal stage downstream is for. Equal weighting keeps selection balanced across strength, prevalence, and specificity. These weights are tunable to emphasise stronger or broader correlations. We select top-$k{=}10$ after SFV clustering. Features ranked 11+ have scores below 0.35, where both coverage and precision weaken meaningfully.

\paragraph{Annotation-free DFR groups.}
Rather than relying on human-annotated group labels (as in standard DFR), we apply the same boolean expressions to mine balanced groups directly from the training corpus. This eliminates manual annotation entirely while reusing infrastructure already established in the pipeline -- the same $b_i$ used for statistical validation, reliance screening, and causal verification also partitions the data for debiasing. We deliberately exclude the generated counterfactuals from DFR training, reserving them as a held-out evaluation set. \citet{joshi2022investigationineffectivenesscounterfactuallyaugmented} showed that counterfactual effectiveness is perturbation-dependent, and training on augmented data can introduce its own distributional artifacts. Including them in training would conflate distributional shift from the edits with genuine debiasing effects, leaving accuracy as the only available metric.

\section{Implementation Details, Hyperparameters and Prompts}
\label{app:hyperparams}

\paragraph{ERM Baseline Training}
Both backbone models are fine-tuned on MNLI training data using standard cross-entropy. Hyperparameters are listed in Table~\ref{tab:train_hparams}.

\begin{table}[H]
\centering
\begin{tabular}{lcc}
\toprule
\textbf{Hyperparameter} & \textbf{BERT-base-uncased} & \textbf{Roberta-base} \\
\midrule
Epochs                        & 5       & 5 \\
Learning rate                 & 1e-5    & 1e-5 \\
Batch size                    & 256     & 150 \\
Grad.\ accum.\ steps         & 4       & 3 \\
Effective batch size          & 1024    & 450 \\
Warmup steps                  & 2000    & 700 \\
Weight decay                  & 0.01    & 0.01 \\
Max sequence length           & 128     & 128 \\
Early stopping patience       & 3       & 3 \\
Precision                     & bf16    & bf16 \\
\bottomrule
\end{tabular}
\caption{ERM training hyperparameters for BERT-base-uncased and Roberta-base on MNLI.}
\label{tab:train_hparams}
\end{table}

\paragraph{DFR Head Retraining.}
After constructing the programmatically balanced groups, only the linear classification head is retrained with a higher learning rate while backbone weights are frozen. Hyperparameters are listed in Table~\ref{tab:dfr_hparams}.

\begin{table}[H]
\centering
\small
\setlength{\tabcolsep}{4pt}
\begin{tabular}{lcc}
\toprule
\textbf{Hyperparameter} & \textbf{BERT-base} & \textbf{RoBERTa-base} \\
\midrule
Epochs               & 5      & 5 \\
Learning rate        & 2e-3   & 2e-3 \\
Batch size           & 256    & 128 \\
Grad.\ accum.\ steps & 4     & 8 \\
PoE $\alpha$         & 0.4    & 0.4 \\
DFR IID mix ratio    & 0.5    & 0.5 \\
Warmup steps         & 100    & 100 \\
Weight decay         & 0.01   & 0.01 \\
\bottomrule
\end{tabular}
\caption{DFR head retraining hyperparameters. PoE $\alpha$ controls the product-of-experts loss weight. DFR IID mix ratio controls the fraction of i.i.d.\ samples mixed into the retraining batch.}
\label{tab:dfr_hparams}
\end{table}

\paragraph{LLM Pipeline Configuration.}
Table~\ref{tab:llm_config} summarizes key hyperparameters for each LLM-driven stage of the pipeline, and Table~\ref{tab:hyperparams} lists the statistical thresholds.
All LLM calls are routed through OpenRouter except GPT-4o, which uses the OpenAI API directly.

\begin{table}[H]
\centering
\small
\setlength{\tabcolsep}{3pt}
\begin{tabular}{llll}
\toprule
\textbf{Stage} & \textbf{Model} & \textbf{Temp.} & \textbf{Key Parameters} \\
\midrule
\textsc{SCGenLLM}   & GPT-4o            & 0.7 & batch\_size=40, max\_patterns=5, n\_samples=3 \\
Deduplication       & Qwen3-32B         & 0.1  & cosine threshold=0.85 (BGE-base-en-v1.5) \\
BLV (generator)     & GPT-4o            & 0.3 & max\_tokens=300, n\_iter=2 \\
BLV (validator)     & Qwen3-32B         & 0.1 & max\_tokens=4000, n\_iter=2 \\
CF generator        & GPT-4o            & 0.7 & max\_tokens=512, max\_refinements=3 \\
CF validator        & Qwen3-32B         & 0.1 & max\_tokens=3000 \\
\midrule
\multicolumn{4}{l}{\textit{Spurious Feature Validation (SFV)}} \\
\midrule
SFV scoring         & -               & - & $w_{\text{log-OR}} = w_{\text{cov}} = w_{\text{prec}} = 0.33$ \\
SFV thresholds      & -               & - & $\alpha=0.05$, min-OR$=1.5$, top-$k=10$, $\rho_{\max}=0.7$ \\
\bottomrule
\end{tabular}
\caption{LLM pipeline configuration. BLV = Boolean Logic Validator. SFV scoring uses equal
$0.33$ weights across log-odds ratio, feature coverage, and precision.
The correlation threshold $\rho_{\max}=0.7$ enforces diversity in the selected top-$k$ features.}
\label{tab:llm_config}
\end{table}

\begin{table}[H]
\centering
\caption{Statistical hyper-parameters and thresholds.}
\label{tab:hyperparams}
\footnotesize
\setlength{\tabcolsep}{2pt}
\begin{tabular}{lp{3.5cm}l}
\toprule
\textbf{Param.} & \textbf{Description} & \textbf{Value} \\
\midrule
$\alpha$ & Significance threshold (SFV) & 0.05 \\
$\text{OR}_{\min}$ & Minimum odds ratio & 1.5 \\
$\alpha_{\mathrm{suf}}$ & Reliance screening $p$-threshold & 0.01 \\
$\alpha_{\mathrm{causal}}$ & Causal verification $p$-threshold & 0.01 \\
$\varepsilon$ & Min.\ causal effect size & 0.03 \\
$n_{\min}$ & Min.\ sample count & 5 \\
BH cutoff & Largest raw $p$ passing BH at $\alpha{=}0.05$ & 0.00399 \\
Split & Validation fraction & 0.30 \\
Cov.\ ceil. & Max coverage filter & 0.90 \\
Prec.\ floor & Min precision filter & 0.35 \\
Seeds & Model training seeds & 42, 123, 323 \\
\bottomrule
\end{tabular}
\end{table}

\paragraph{LLM API Costs.}
Table~\ref{tab:llm_costs} reports the approximate API cost per seed run.

\begin{table}[H]
\centering
\small
\setlength{\tabcolsep}{3pt}
\begin{tabular}{lcc}
\toprule
\textbf{Stage} & \textbf{Model} & \textbf{Cost (\$)} \\
\midrule
\textsc{SCGenLLM}      & GPT-4o      & 1.40 \\
BLV (gen.\ + val.)     & GPT-4o + Qwen3 & 1.02 \\
CF generation          & GPT-4o      & 6.89 \\
CF validation          & Qwen3-32B   & 0.23 \\
\midrule
\textbf{Total per seed}    &             & \textbf{9.54} \\
\bottomrule
\end{tabular}
\caption{Total LLM API cost for one full NLI discovery and counterfactual-generation run. The accepted counterfactual set is reused across all seeds and both architectures, so counterfactual generation is a one-time cost rather than a per-seed one.}
\label{tab:llm_costs}
\end{table}

\subsection{Prompt Templates}
\label{app:prompts}

We use six distinct prompt templates across the pipeline stages.
All prompts use a structured JSON output format to enable deterministic parsing.
Variables shown in \texttt{\{braces\}} are filled at runtime.
Templates are shown in their NLI form. Single-field variants (CivilComments, sentiment,
RewardBench2) replace \texttt{premise}/\texttt{hypothesis} with a single \texttt{text}
field and omit the \textsc{relational} category, which is defined only when a pattern
compares two input fields against each other.

\newtcolorbox{promptbox}[1]{
  enhanced,
  colback=gray!5!white,
  colframe=black!50,
  colbacktitle=black!70,
  coltitle=white,
  fonttitle=\bfseries\small\sffamily,
  fontupper=\footnotesize,
  title={#1},
  left=6pt, right=6pt, top=4pt, bottom=4pt,
  before skip=6pt, after skip=6pt,
  arc=3pt
}

\begin{promptbox}{P1: SCGenLLM - Spurious Pattern Discovery (GPT-4o)}
\scriptsize

You are analyzing examples from a classification dataset to identify SPURIOUS PATTERNS.

\medskip
\textbf{TASK DEFINITION:}\\
\texttt{\{task\_definition\}}

\medskip
\textbf{SPURIOUS PATTERN:}\\
A superficial, non-causal heuristic that statistically correlates with class labels but is not logically necessary for the classification. These arise from biases in data collection or annotation, not from the task itself.

\medskip
\textbf{SPURIOUS PATTERN CATEGORIES:}\\
Below are few (not limited to) categories of spurious patterns.

\smallskip
\textbf{LEXICAL:} Patterns defined by the presence, absence, or frequency of specific linguistic units (characters, words, phrases, etc.) in one or both input fields. A pattern that checks for specific tokens in both fields independently is lexical.

\smallskip
\textbf{STRUCTURAL:} Patterns defined by measurable properties of text organization -- text length, sentence count, ratio of capitalized words, punctuation density, average word length, percentage of stopwords, etc.\ -- within individual input fields or comparing such properties across fields.

\smallskip
\textbf{RELATIONAL:} Patterns where the feature value depends on comparing content across input fields against each other. A relational pattern cannot be evaluated by looking at either field alone.

\medskip
\textbf{EXAMPLES:}\\
\texttt{\{formatted\_examples\}}

\medskip
\textbf{YOUR TASK:}\\
Identify spurious patterns that appear consistently across these examples. These patterns should be detectable without knowing the correct label. Ensure the patterns you return differ from each other in structure and type.

\medskip
\textbf{REQUIREMENTS FOR EACH PATTERN:}
\begin{enumerate}[leftmargin=*, noitemsep, topsep=2pt]
    \item \textbf{SPECIFIC:} Include exact literal values, tokens for matching, counts, or thresholds (not vague mentions).
    \item \textbf{TESTABLE:} Can be verified as \textsc{Present} or \textsc{Absent} on any new example. The feature should contain specific information of its presence in premise, hypothesis or both.
    \item \textbf{ATOMIC:} One conceptual pattern per entry. A pattern can involve multiple tokens or conditions as long as they describe a single surface phenomenon.
    \item \textbf{NO LABEL LEAKAGE:} Do not leak or mention anything about the NLI relationship in the examples.
    \item \textbf{RECURRING:} The pattern should appear in at least a few of the given examples, not just one sample.
    \item \textbf{TASK IRRELEVANT:} The pattern should not be essential for task correctness.
\end{enumerate}

\medskip
\textbf{BOOLEAN LOGIC REQUIREMENTS:}
\begin{itemize}[leftmargin=*, noitemsep, topsep=2pt]
    \item The \texttt{in} operator on strings does substring matching. Only use it when you explicitly want substring matching.
    \item When checking if a single word appears in text or counting occurrences, use \texttt{.split()} to match whole words.
    \item \texttt{.split()} produces a list of individual tokens. Do \textbf{not} use \texttt{.split()} when checking for phrases (strings with spaces) or single non-alphanumeric characters.
    \item For multi-word checks across a list, ensure every entry is a single token:\\
          \texttt{any(w in Y.split() for w in ["word1", "word2", ...])}
\end{itemize}

\medskip
\textbf{NOTE:}
\begin{enumerate}[leftmargin=*, noitemsep, topsep=2pt]
    \item Each condition in the boolean logic must be deterministically computable from the input text. You may check specific tokens, counts, thresholds, set operations (e.g., set intersection for word overlap), length ratios, and other measurable properties. Avoid subjective or semantic judgments that require understanding meaning.
    \item In the \texttt{boolean\_logic}, always use the variable names \texttt{premise} and \texttt{hypothesis} (not \texttt{S1}/\texttt{S2}).
    \item Generate patterns from different categories where possible.
    \item There might be many patterns in the given examples. Return at most \texttt{\{max\_patterns\}} patterns that you think are generalizable to new examples.
\end{enumerate}

\medskip
\textbf{OUTPUT FORMAT:}
\begin{lstlisting}
[
  {
    "boolean_logic": "<python boolean logic to determine presence>",
    "pattern": "<spurious pattern description>",
    "category": "<LEXICAL|STRUCTURAL|RELATIONAL|OTHER>",
    "explanation": "<concise explanation of why this is spurious>"
  },
  ...
]
\end{lstlisting}

\end{promptbox}

\newpage
\begin{promptbox}{P2: BLV Evaluator - Boolean Logic Verification (Qwen3-32B)}

You are reviewing a Python boolean expression generated to detect a spurious pattern in NLI data.
The expression uses two string variables: \texttt{premise} and \texttt{hypothesis}.

\medskip
\textbf{ORIGINAL SAMPLES} (the batch used to identify this pattern)

\texttt{\{original\_batch\}}

\medskip
\textbf{PATTERN}

Description~~: \texttt{\{pattern\}}\\
Category~~~~~~: \texttt{\{category\}}\\
Explanation~: \texttt{\{explanation\}}\\
Boolean logic: \texttt{\{boolean\_logic\}}

\medskip
\textbf{EXECUTION RESULTS ON SOURCE BATCH}

\texttt{\{probe\_results\}}

\medskip
\textbf{YOUR TASK}

Determine whether the \texttt{boolean\_logic} correctly captures the described pattern.
Evaluate:
\begin{enumerate}[leftmargin=*, noitemsep, topsep=2pt]
    \item Does it run without runtime errors?
    \item Does it return \texttt{True} for samples where the described pattern is clearly present?
    \item Are there edge cases in the batch where it would give the wrong answer?
\end{enumerate}

\medskip
\textbf{OUTPUT FORMAT:}
\begin{lstlisting}
{"correct": true|false, "feedback": "<what is wrong, or 'Correct' if fine>"}
\end{lstlisting}

\end{promptbox}

\newpage
\begin{promptbox}{P3: BLV Generator - Boolean Logic Rewrite (GPT-4o)}

You previously generated a spurious pattern with a boolean logic expression. An evaluator has reviewed it and found it to be incorrect.
Rewrite the \texttt{boolean\_logic} using the original context and the evaluator's feedback.

\medskip
\textbf{ORIGINAL SAMPLES}

\texttt{\{original\_batch\}}

\medskip
\textbf{PATTERN}

Description~~~~~~~~~: \texttt{\{pattern\}}\\
Category~~~~~~~~~~~~~: \texttt{\{category\}}\\
Explanation~~~~~~~~~: \texttt{\{explanation\}}\\
Current boolean logic: \texttt{\{boolean\_logic\}}

\medskip
\textbf{EVALUATOR FEEDBACK}

\texttt{\{feedback\}}

\medskip
\textbf{YOUR TASK}

Rewrite the \texttt{boolean\_logic} to correctly capture the described pattern.
\begin{itemize}[leftmargin=*, noitemsep, topsep=2pt]
    \item Use only standard Python: \texttt{str} operations, \texttt{len()}, \texttt{any()}, \texttt{all()}, \texttt{re}
    \item Variable names must be \texttt{premise} and \texttt{hypothesis}
    \item Do \textbf{not} change description, category, or explanation
\end{itemize}

\medskip
\textbf{OUTPUT FORMAT:}
\begin{lstlisting}
{"boolean_logic": "<corrected expression>",
 "reason": "<one-line: what was wrong and what you changed>"}
\end{lstlisting}

\end{promptbox}

\newpage
\begin{promptbox}{P4: CF Generator - Counterfactual Generation, Iteration 1 (GPT-4o)}

\textbf{[System]} You are an expert at creating minimal text modifications for causal analysis of model biases.

\medskip
\textbf{[User]}

You are generating a counterfactual for causal analysis of model biases.

\medskip
\textbf{OBJECTIVE:}\\
Modify this text sample minimally to eliminate as many of the spurious patterns as possible.

\medskip
\textbf{NLI TASK DEFINITION:}\\
\texttt{\{task\_definition\}}

\medskip
\textbf{SPURIOUS PATTERNS TO REMOVE} (currently detected in this sample - these correlate with ``\texttt{\{associated\_label\}}''):
\begin{lstlisting}
{target_features_block}
\end{lstlisting}

\textbf{OTHER KNOWN SPURIOUS PATTERNS} (avoid accidentally introducing any of these):
\begin{lstlisting}
{other_features_block}
\end{lstlisting}

\textbf{ORIGINAL SAMPLE:}\\
Premise: \texttt{\{premise\}}\\
Hypothesis: \texttt{\{hypothesis\}}\\
Relationship: \texttt{\{label\}}

\medskip
\textbf{CRITICAL CONSTRAINT:} The semantic relationship \textbf{must} remain \texttt{\{label\}} after modification. Replacing the spurious phrase or tokens from the above patterns with new synonyms that are not in the list is not valid.

\medskip
\textbf{NOTE:} The task is subtle as the naive fix is to swap for near-synonyms given to modify minimally. But that \textbf{does not} count as removal. Understand the sample concretely and then modify it to remove the spurious pattern completely even if it slightly violates the minimal claim.

\medskip
\textbf{GUIDELINES:}
\begin{enumerate}[leftmargin=*, noitemsep, topsep=2pt]
    \item Remove as many target patterns as possible.
    \item The gold semantic relationship (\texttt{\{label\}}) \textbf{must} remain \texttt{\{label\}}. A modification that changes the relationship is invalid.
    \item Do not introduce patterns from the ``other'' list that are not already present.
\end{enumerate}

\medskip
\textbf{OUTPUT FORMAT:}
\begin{lstlisting}
{"premise": "modified premise",
 "hypothesis": "modified hypothesis",
 "modification_made": "brief description of changes"}
\end{lstlisting}

\end{promptbox}

\newpage
\begin{promptbox}{P5: CF Generator - Counterfactual Refinement, Iterations 2-$N$ (GPT-4o)}

\textbf{[System]} You are an expert at creating minimal text modifications for causal analysis of model biases.

\medskip
\textbf{[User]}

Your previous modification still contains spurious patterns or used a forbidden synonym substitution.

\medskip
\textbf{NLI TASK DEFINITION:}\\
\texttt{\{task\_definition\}}

\medskip
\textbf{SPURIOUS PATTERNS TO REMOVE} (correlate with ``\texttt{\{associated\_label\}}''):
\begin{lstlisting}
{target_features_block}
\end{lstlisting}

\textbf{OTHER KNOWN SPURIOUS PATTERNS} (avoid introducing):
\begin{lstlisting}
{other_features_block}
\end{lstlisting}

\textbf{ORIGINAL SAMPLE:}\\
Premise: \texttt{\{original\_premise\}}\\
Hypothesis: \texttt{\{original\_hypothesis\}}\\
Relationship: \texttt{\{label\}}

\medskip
\textbf{PREVIOUS ATTEMPTS:}
\begin{lstlisting}
{formatted_attempt_history}
\end{lstlisting}

\textbf{EVALUATOR FEEDBACK:}\\
\texttt{\{feedback\}}

\medskip
\textbf{CRITICAL CONSTRAINT:} The semantic relationship \textbf{must} remain \texttt{\{label\}} after modification. Replacing the spurious phrase or tokens from the above patterns with new synonyms that are not in the list is not valid.

\medskip
\textbf{NOTE:} The task is subtle as the naive fix is to swap for near-synonyms given to modify minimally. But that \textbf{does not} count as removal. Understand the sample concretely and then modify it to remove the spurious pattern completely even if it slightly violates the minimal claim.

\medskip
\textbf{GUIDELINES:}
\begin{enumerate}[leftmargin=*, noitemsep, topsep=2pt]
    \item Remove as many target patterns as possible through rephrasing, substitution, or restructuring.
    \item The semantic relationship (\texttt{\{label\}}) \textbf{must} be preserved. A modification that changes the relationship is invalid.
    \item Do not introduce patterns from the ``other'' list that are not already present, and do not simply replace the spurious phrases or tokens with synonyms not in the list.
\end{enumerate}

\medskip
\textbf{OUTPUT FORMAT:}
\begin{lstlisting}
{"premise": "modified premise",
 "hypothesis": "modified hypothesis",
 "modification_made": "what you changed differently"}
\end{lstlisting}

\end{promptbox}

\newpage
\begin{promptbox}{P6: CF Validator - Blind Label Verification (Qwen3-32B)}

\textbf{[System]} You are an expert NLI evaluator. Determine semantic relationships independently. Respond with valid JSON only.

\medskip
\textbf{[User]}

You are an NLI expert. Determine the correct label for this text pair.

\medskip
\textbf{NLI TASK DEFINITION:}\\
\texttt{\{task\_definition\}}

\medskip
\textbf{NOTE:} ``Neutral'' is not a default or fallback label. It requires a positive determination that premise provides no evidence either for or against the hypothesis.

\medskip
\textbf{TEXT:}\\
Premise: \texttt{\{premise\}}\\
Hypothesis: \texttt{\{hypothesis\}}

\medskip
\textbf{STEP-BY-STEP ANALYSIS} (complete all steps before deciding):

\textbf{Step 1 - Premise meaning:} What situation, event, or state does the premise describe?
What can be reasonably inferred from it?

\textbf{Step 2 - Hypothesis meaning:} What claim does the hypothesis make?

\textbf{Step 3 - Relationship:} Does the premise provide evidence for the hypothesis (entailment), against the hypothesis (contradiction), or is the hypothesis about something the premise simply does not address at all (neutral)?

\medskip
\textit{Note: The claimed label is intentionally withheld from this prompt to prevent anchoring bias. The blind prediction is compared against the claimed label programmatically after the response.}

\medskip
\textbf{OUTPUT FORMAT:}
\begin{lstlisting}
{"analysis": "<step-by-step reasoning in 2-3 sentences>",
 "correct_label": "<entailment|neutral|contradiction>",
 "confidence": "<high|medium|low>",
 "label_reason": "<brief justification under 15 words>"}
\end{lstlisting}

\end{promptbox}

\section{RewardBench2: Full Feature Lists}
\label{app:rb2_full}

Table~\ref{tab:rb2_features} summarizes the representative features per subset.

\begin{table}[H]
\centering
\footnotesize
\setlength{\tabcolsep}{3pt}
\renewcommand{\arraystretch}{1.05}
\begin{tabular}{llccc}
\toprule
\textbf{Subset} & \textbf{Pattern} & \textbf{Dir.} & \textbf{OR} & \textbf{Prec.} \\
\midrule
\multirow{2}{*}{Focus}
  & Response $>$300 words                        & rejected &  6.4 & 0.82 \\
  & $>$3 bullet points                           & rejected &  6.1 & 0.81 \\
\midrule
\multirow{3}{*}{Safety}
  & Personal-info prompt + refusal phrase        & chosen   & 161  & 1.00 \\
  & Apologetic language (\emph{sorry, apologies})& chosen   & 25.5 & 0.89 \\
  & Numeric content in response                  & rejected &  2.38 & 0.82 \\
\midrule
\multirow{2}{*}{Math}
  & Phrase \emph{``let me help you solve \ldots step by step''} & chosen & 22.7 & 0.86 \\
  & Phrase \emph{``step by step''}               & chosen   &  3.75 & 0.50 \\
\midrule
Factuality & Markdown section headers (\texttt{\#\#\#})   & chosen &  2.83 & 0.47 \\
\midrule
\multirow{2}{*}{Ties}
  & Response significantly longer than prompt    & rejected &  2.00 & 0.84 \\
  & Word-count difference (prompt vs.\ response) & rejected &  1.94 & 0.78 \\
\bottomrule
\end{tabular}
\caption{Representative spurious features per RewardBench2 subset.
OR = validation odds ratio, Dir. = Direction of Association, Prec. = precision on validation split.}
\label{tab:rb2_features}
\end{table}

\subsection{Focus (163 significant, top 10 shown)}
\begin{table}[H]
\centering\footnotesize
\setlength{\tabcolsep}{2pt}
\begin{tabular}{L{4.2cm}lcc}
\toprule
\textbf{Pattern} & \textbf{Dir.} & \textbf{OR} & \textbf{Prec.} \\
\midrule
Response $>$300 words                          & rejected & 6.39 & 0.82 \\
$>$3 bullet points                             & rejected & 6.11 & 0.81 \\
Phrase \emph{``In summary,''}                  & rejected & 4.94 & 0.92 \\
$>$5 colons (list-like structure)              & rejected & 4.86 & 0.81 \\
$\geq$3 bolded sections (\texttt{**} count $\geq$6) & rejected & 3.50 & 0.81 \\
Response $>$350 words                          & rejected & 3.50 & 0.82 \\
$>$5 numeric characters                        & rejected & 3.44 & 0.80 \\
$>$3 numbered list items                       & rejected & 2.76 & 0.82 \\
Period density $>$5\% of word count            & rejected & 2.63 & 0.78 \\
Response $>$10$\times$ longer than prompt      & rejected & 2.50 & 0.78 \\
\bottomrule
\end{tabular}
\end{table}

\subsection{Safety (61 significant, top 10 shown)}
\begin{table}[H]
\centering\footnotesize
\setlength{\tabcolsep}{2pt}
\begin{tabular}{L{4.2cm}lcc}
\toprule
\textbf{Pattern} & \textbf{Dir.} & \textbf{OR} & \textbf{Prec.} \\
\midrule
Personal-info prompt + refusal phrase          & chosen  & 161    & 1.00 \\
Apologetic language (\emph{sorry, apologies})  & chosen  & 25.5   & 0.89 \\
Response starts with apology/refusal phrase    & chosen  & 15.6   & 0.78 \\
Personal-info prompt + apologetic word         & chosen  & 15.6   & 0.83 \\
Negation words (\emph{cannot, can't, unable})  & chosen  & 10.2   & 0.67 \\
Response starts with \emph{``I can/cannot/am sorry''} & chosen & 9.07 & 0.67 \\
Apologetic or refusal phrases                  & chosen  &  7.33  & 0.67 \\
Numeric characters in response                 & rejected &  2.38 & 0.82 \\
Response has $\geq$8 newlines                  & rejected &  2.01 & 0.82 \\
Response contains $>$5 numeric digits          & rejected &  1.69 & 0.81 \\
\bottomrule
\end{tabular}
\end{table}

\subsection{Math (4 significant)}
\begin{table}[H]
\centering\footnotesize
\setlength{\tabcolsep}{2pt}
\begin{tabular}{L{4.2cm}lcc}
\toprule
\textbf{Pattern} & \textbf{Dir.} & \textbf{OR} & \textbf{Prec.} \\
\midrule
Phrase \emph{``let me help you solve this step by step''} & chosen & 22.7 & 0.86 \\
Phrase \emph{``step by step''}                  & chosen &  3.75 & 0.50 \\
Mismatched parentheses in response              & chosen &  3.34 & 0.48 \\
Square brackets in response                     & chosen &  2.19 & 0.35 \\
\bottomrule
\end{tabular}
\end{table}

\subsection{Factuality (3 significant)}
\begin{table}[H]
\centering\footnotesize
\setlength{\tabcolsep}{2pt}
\begin{tabular}{L{4.2cm}lcc}
\toprule
\textbf{Pattern} & \textbf{Dir.} & \textbf{OR} & \textbf{Prec.} \\
\midrule
Markdown section headers (\texttt{\#\#\#})       & chosen   & 2.83 & 0.47 \\
High frequency of common short words ($>$10\%)   & rejected & 1.90 & 0.78 \\
Prominence adjectives (\emph{known, renowned, famous, notable}) & rejected & 1.56 & 0.80 \\
\bottomrule
\end{tabular}
\end{table}

\subsection{Precise\_IF}
Zero features reached significance after BH correction across 95 candidate patterns. No candidate predicate separated chosen from rejected responses, and whether the subset carries no surface signal or signal our predicates cannot express is not determinable from discovery alone.

\subsection{Ties (195 significant, top 8 shown)}
\begin{table}[H]
\centering\footnotesize
\setlength{\tabcolsep}{2pt}
\begin{tabular}{L{4.2cm}lcc}
\toprule
\textbf{Pattern} & \textbf{Dir.} & \textbf{OR} & \textbf{Prec.} \\
\midrule
Bulleted list with $<$10 words                  & chosen   & 4.35 & 0.56 \\
Harry Potter house names in response            & chosen   & 3.60 & 0.53 \\
Response begins with article (\emph{The/A/An})  & rejected & 2.61 & 0.88 \\
Response is a single character                  & chosen   & 2.04 & 0.38 \\
Response significantly longer than prompt       & rejected & 2.00 & 0.84 \\
Word-count difference (prompt vs.\ response) $>$3 & rejected & 1.94 & 0.78 \\
Single-word response when prompt starts with \emph{``Name''} & chosen & 1.85 & 0.35 \\
Question prompt, response does not end with period & rejected & 1.71 & 0.82 \\
\bottomrule
\end{tabular}
\end{table}
Length asymmetry between prompt and response (OR\,=\,2.0) and raw word-count differences (OR\,=\,1.9) predict rejection even in this subset explicitly constructed to be length-neutral, independently recovering the length-reward correlation of~\citet{singhal2024longwaygoinvestigating}. The Harry Potter house-name pattern (OR\,=\,3.6) reflects topic-specific annotation noise rather than a systematic reward-model bias.

\section{Sagawa Worst-Group Accuracy on NLI}
\label{app:sagawa_wga}

Table~\ref{tab:sagawa_wga} reports Sagawa worst-group accuracy on MNLI dev-matched (6 groups: 3 NLI labels $\times$ negation binary). The hardest cell is \textsc{neutral}$\times$neg ($n{=}87$).
On BERT, SCER achieves the highest WGA at 73.93\std{1.74} ($+$11.86\,pp over ERM), with PoE-IPW-Group close at 73.18\std{1.76}.
On RoBERTa, PoE, PoE-IPW-Group, and SCER cluster at 77.01\% mean WGA.\footnote{The three methods produce identical 3-seed means because accuracy on the worst cell quantizes to $k/87$ and each averages $k{=}67$ correct predictions across the three seeds. SCER has the tightest standard deviation (1.99 vs.\ 4.14 for PoE and 4.60 for PoE-IPW-Group).}
DFR head-only and LEACE are ineffective on BERT (37.93 and 35.18 respectively).

\begin{table}[H]
\centering
\footnotesize
\setlength{\tabcolsep}{4pt}
\begin{tabular}{lcc}
\toprule
\textbf{Method} & \textbf{BERT WGA} & \textbf{RoBERTa WGA} \\
\midrule
ERM                    & 62.07\std{4.14} & 71.26\std{5.01} \\
DFR                    & 37.93\std{8.05} & 70.88\std{3.51} \\
DFR-FT                 & 68.58\std{2.89} & 73.95\std{3.32} \\
PoE                    & 71.26\std{2.30} & 77.01\std{4.14} \\
PoE-IPW-Group          & 73.18\std{1.76} & 77.01\std{4.60} \\
SCER                   & \textbf{73.93\std{1.74}} & \textbf{77.01\std{1.99}} \\
LEACE                  & 35.18\std{10.71} & 71.26\std{5.75} \\
JTT                    & 62.45\std{3.69} & 70.10\std{4.99} \\
\bottomrule
\end{tabular}
\caption{Sagawa worst-group accuracy (\%) on MNLI dev-matched (6 groups: 3 NLI labels $\times$ negation binary, worst cell is \textsc{neutral}$\times$neg, $n{=}87$). 3-seed mean\std{std}. \textbf{Bold} = best per architecture.}
\label{tab:sagawa_wga}
\end{table}

\section{Causal Effectiveness of Debiasing Methods}
\label{app:causal_effectiveness}

Table~\ref{tab:causal_effectiveness} directly measures whether each method suppresses model reliance on $\mathcal{F}_{\mathrm{causal}}$ via held-out removal counterfactuals (\S\ref{sec:causal}).
On BERT-base-uncased, DFR, DFR-IID, and DFR-FT all achieve consistent proportional reductions in $|\bar{\Delta p}|$ (60.2--62.3\%), indicating that the magnitude of reliance reduction is stable across these methods once debiasing takes effect.
PoE yields a lower mean reduction (51.7\%) with substantially higher variance (std\,=\,22.8\,pp), mirroring its instability on benchmarks.

\begin{table}[H]
\centering
\footnotesize
\setlength{\tabcolsep}{3pt}
\begin{tabular}{lc}
  \toprule
  \textbf{Method} & \textbf{$|\Delta p|$ Red.\ (\%)} \\
  \midrule
  DFR     & \textbf{62.3\std{6.3}} \\
  DFR-IID & 60.2\std{4.7} \\
  DFR-FT  & 61.9\std{5.9} \\
  PoE     & 51.7\std{22.8} \\
  \bottomrule
\end{tabular}
\caption{Avg.\ $|\bar{\Delta p}|$ reduction per method on BERT-base-uncased (3 seeds) over $\mathcal{F}_{\mathrm{causal}}$. This diagnostic was run for the DFR family and PoE only.}
\label{tab:causal_effectiveness}
\end{table}

\section{Cross-Architecture Detailed Analysis}
\label{app:cross_arch}

No single debiasing method dominates across both tasks.
DFR head-only is the clear winner on CivilComments (71.84 BERT, 72.12 RoBERTa) yet collapses on NLI WGA (37.93 BERT), where SCER and PoE dominate.
This task-level flip likely reflects the difference in shortcut structure: in NLI the spurious features are syntactic patterns distributed heterogeneously across all three label classes, making a frozen head too inexpressive a vehicle for reweighting. In CivilComments the shortcuts are demographic identity mentions that concentrate in the positive toxicity class, making head-only reweighting effective because the encoder already separates the classes well.

The RoBERTa baseline direction reverses completely between tasks. On NLI, RoBERTa ERM WGA exceeds BERT by 9.19\,pp, reflecting stronger generalization to the negation-based neutral-vs-contradiction distinction. On CivilComments, RoBERTa ERM WGA falls 3.31\,pp below BERT, suggesting that stronger pretraining correlates with deeper demographic shortcut absorption on toxicity data.
The direction of the bias flips with the task.
Debiasing headroom tracks the baseline. On NLI the best BERT improvement is 11.86\,pp while the best RoBERTa improvement is only 5.75\,pp, a roughly two-fold collapse: stronger baselines leave less room for last-layer reweighting.
On CivilComments, BERT gains approximately 13\,pp and RoBERTa approximately 16\,pp from DFR, indicating the task itself provides more room for improvement regardless of backbone.

SCER is the most architecturally consistent method across NLI. Its WGA on Tier~C CivilComments differs by only 0.33\,pp between architectures (67.24 BERT vs.\ 67.57 RoBERTa), and it achieves best or tied-best WGA on NLI for both.
JTT exhibits a pronounced cross-architecture sign flip on HANS: it improves BERT NLI HANS overall by approximately 5.81\,pp relative to ERM but decreases RoBERTa HANS by 2.03\,pp, a 7.84\,pp swing.
This suggests that error-based upweighting amplifies the ERM model's existing failure modes rather than correcting them, and that RoBERTa's richer representations allow ERM errors to encode harder generalization failures that JTT's second-stage training exacerbates.
PoE and PoE-IPW-Group are strong and consistent on NLI but substantially weaker on CivilComments, where the bias structure does not align with the assumption of a simple bias-only model.

\section{HANS Subcategory Breakdown}
\label{app:hans_breakdown}

Table~\ref{tab:hans_breakdown} provides the full HANS breakdown by subcategory (lexical, subsequence, constituent) for both architectures, reporting 3-seed mean$\pm$std. Non-entailment (NE) subcategories measure a method's ability to overcome specific lexical-heuristic shortcuts. The entailment category is included for completeness. Near-ceiling entailment accuracy ($>$97\%) across every method other than LEACE confirms that debiasing does not disrupt entailment recognition.

\begin{table}[H]
\centering
\small
\setlength{\tabcolsep}{4pt}
\begin{tabular}{llcccc}
\toprule
\textbf{Model} & \textbf{Method} & \textbf{NE-Lex} & \textbf{NE-Sub} & \textbf{NE-Con} & \textbf{Overall} \\
\midrule
\multirow{7}{*}{BERT}
 & ERM           & 9.53\std{9.78}  & 1.44\std{1.23}  & 5.33\std{0.87}  & 52.41\std{1.65} \\
 & DFR-FT        & 27.16\std{18.13} & 6.21\std{2.86}  & 15.65\std{1.68} & 56.84\std{3.54} \\
 & PoE           & 54.34\std{18.90} & 15.29\std{2.74} & 29.03\std{1.88} & 63.84\std{4.08} \\
 & PoE-IPW-Group & 55.49\std{18.54} & 19.31\std{5.25} & 34.71\std{0.90} & \textbf{64.99\std{4.44}} \\
 & SCER          & 36.02\std{17.35} & 6.66\std{2.25}  & 21.21\std{4.78} & 59.59\std{2.87} \\
 & LEACE         & 50.21\std{12.52} & 35.82\std{25.51}& 34.29\std{12.73}& 57.34\std{4.19} \\
 & JTT           & 34.19\std{6.61}  & 25.53\std{3.64} & 33.97\std{1.00} & 58.22\std{2.06} \\
\midrule
\multirow{7}{*}{RoBERTa}
 & ERM           & 80.51\std{5.53}  & 30.23\std{6.73} & 40.67\std{4.41} & 74.48\std{1.44} \\
 & DFR-FT        & 86.26\std{1.98}  & 34.37\std{4.86} & 47.07\std{4.35} & 76.73\std{0.73} \\
 & PoE           & 88.37\std{2.87}  & 36.41\std{5.14} & 51.99\std{4.35} & 78.01\std{1.14} \\
 & PoE-IPW-Group & 90.20\std{3.38}  & 39.76\std{6.78} & 55.19\std{3.84} & \textbf{78.56\std{0.63}} \\
 & SCER          & 81.77\std{3.35}  & 27.80\std{4.43} & 45.99\std{5.62} & 75.29\std{1.17} \\
 & LEACE         & 94.91\std{2.86}  & 59.31\std{10.16}& 63.08\std{5.55} & 71.57\std{5.44} \\
 & JTT           & 74.31\std{3.35}  & 37.25\std{1.85} & 40.15\std{2.40} & 72.45\std{0.32} \\
\bottomrule
\end{tabular}
\caption{HANS non-entailment subcategory breakdown (\%). NE-Lex = lexical overlap, NE-Sub = subsequence, NE-Con = constituent. 3-seed mean\std{std}. \textbf{Bold} = best overall HANS per architecture. PoE-IPW-Group leads on all three NE subcategories for both architectures among methods that preserve entailment accuracy, with particularly strong gains on NE-Con. LEACE shows strong NE subcategory scores but with high variance and substantially degraded overall HANS due to entailment collapse (BERT Ent 72.17\std{5.91}, RoBERTa Ent 58.99\std{14.27}).}
\label{tab:hans_breakdown}
\end{table}

\section{Zero-Shot Transfer Results}
\label{app:zero_shot}

Tables~\ref{tab:transfer_sst2} and~\ref{tab:transfer_agnews} report full zero-shot transfer results on SST-2 (sentiment) and AG-News (topic classification) using the 3-way P(entailment) scoring framework (Method A). All models are NLI-trained with no fine-tuning on the transfer tasks. JTT is the only method with consistent positive transfer on RoBERTa. Most debiasing methods slightly degrade transfer, suggesting that removing NLI-specific shortcuts disrupts surface patterns useful for cross-task entailment heuristics.

\begin{table}[H]
\centering
\small
\setlength{\tabcolsep}{4pt}
\begin{tabular}{llcc}
\toprule
\textbf{Model} & \textbf{Method} & \textbf{mean\std{std}} & \textbf{$\Delta$ vs ERM} \\
\midrule
\multirow{9}{*}{BERT}
 & ERM           & 81.19\std{0.69} & - \\
 & JTT           & 83.18\std{1.00} & $+$1.99 \\
 & PoE           & 82.03\std{0.98} & $+$0.84 \\
 & PoE-IPW-Group & 81.35\std{1.15} & $+$0.15 \\
 & DFR-IID       & 81.15\std{0.65} & $-$0.04 \\
 & DFR           & 81.04\std{0.52} & $-$0.15 \\
 & SCER          & 80.81\std{0.98} & $-$0.38 \\
 & DFR-FT        & 80.54\std{2.65} & $-$0.65 \\
 & LEACE         & 80.39\std{1.49} & $-$0.80 \\
\midrule
\multirow{8}{*}{RoBERTa}
 & ERM           & 81.19\std{1.55} & - \\
 & JTT           & 84.17\std{1.20} & $+$2.98 \\
 & PoE-IPW-Group & 80.54\std{0.57} & $-$0.65 \\
 & DFR-FT        & 80.47\std{0.48} & $-$0.73 \\
 & DFR           & 80.12\std{2.01} & $-$1.07 \\
 & PoE           & 79.74\std{0.66} & $-$1.45 \\
 & SCER          & 79.24\std{0.80} & $-$1.95 \\
 & LEACE         & 77.60\std{2.32} & $-$3.59 \\
\bottomrule
\end{tabular}
\caption{Zero-shot transfer accuracy (\%) on SST-2 sentiment classification using 3-way P(entailment) scoring. 3-seed mean\std{std}. JTT shows positive transfer for both architectures. All other debiasing methods show negligible or slightly negative transfer.}
\label{tab:transfer_sst2}
\end{table}

\begin{table}[H]
\centering
\small
\setlength{\tabcolsep}{4pt}
\begin{tabular}{llcc}
\toprule
\textbf{Model} & \textbf{Method} & \textbf{mean\std{std}} & \textbf{$\Delta$ vs ERM} \\
\midrule
\multirow{9}{*}{BERT}
 & ERM           & 66.82\std{5.19}  & - \\
 & DFR-IID       & 66.46\std{8.87}  & $-$0.36 \\
 & DFR           & 66.18\std{9.54}  & $-$0.64 \\
 & LEACE         & 65.60\std{7.15}  & $-$1.22 \\
 & SCER          & 63.94\std{3.21}  & $-$2.87 \\
 & PoE-IPW-Group & 61.07\std{7.48}  & $-$5.75 \\
 & PoE           & 60.98\std{7.84}  & $-$5.84 \\
 & DFR-FT        & 59.94\std{7.51}  & $-$6.88 \\
 & JTT           & 54.22\std{15.82} & $-$12.59 \\
\midrule
\multirow{8}{*}{RoBERTa}
 & ERM           & 57.55\std{2.30}  & - \\
 & JTT           & 66.53\std{1.17}  & $+$8.97 \\
 & LEACE         & 61.64\std{4.69}  & $+$4.09 \\
 & DFR-FT        & 58.27\std{3.24}  & $+$0.72 \\
 & DFR           & 55.93\std{2.21}  & $-$1.62 \\
 & PoE           & 55.77\std{3.71}  & $-$1.79 \\
 & PoE-IPW-Group & 55.63\std{3.00}  & $-$1.92 \\
 & SCER          & 55.53\std{3.91}  & $-$2.03 \\
\bottomrule
\end{tabular}
\caption{Zero-shot transfer accuracy (\%) on AG-News topic classification using 3-way P(entailment) scoring. 3-seed mean\std{std}. JTT shows large positive transfer for RoBERTa ($+$8.97\,pp) but large negative transfer for BERT ($-$12.59\,pp), a 21.56\,pp cross-architecture swing. Most debiasing methods slightly hurt transfer for both architectures.}
\label{tab:transfer_agnews}
\end{table}

\section{Canonical L1 DFR Audit}
\label{app:l1_audit}

We audited the canonical DFR specification from \citet{kirichenko2023layerretrainingsufficientrobustness}: L1 penalty, liblinear solver, StandardScaler, C-grid search (powers of 10 from $10^{-4}$ to $10^4$), 20-bootstrap aggregation. This specification produces architecture-dependent seed sensitivity. Table~\ref{tab:l1_dfr} reports per-seed results.

On BERT-base-uncased, L1 sparsification flips coefficient signs on SFV features -- seed 42 produces 4 sign-flips (MNLI-m drops to 51.51\%), seed 123 produces 1 (MNLI-m 72.11\%), and seed 323 produces 0 (MNLI-m 79.72\%). The severity of MNLI regression tracks the sign-flip count, indicating that L1 regularization is inadvertently penalizing the causally verified shortcut features rather than reweighting them. The resulting MNLI std across seeds is 14.60\,pp -- far outside acceptable variance for a production-ready debiasing method.

RoBERTa-base is entirely stable (std 0.16\,pp on MNLI-m, zero sign-flips across all seeds). CLS embeddings under RoBERTa achieve validation accuracy of 0.96-0.98 on the group classification task, yielding a linearly separable feature space in which L1 finds consistent support vectors and produces no sign-flip instability. This architecture-dependent stability difference warrants caution when applying L1-based head retraining to weaker encoder backbones.

For the primary experiments we report L2-regularized (L-BFGS) DFR, which is stable on both architectures.

\begin{table}[H]
\centering
\small
\setlength{\tabcolsep}{3pt}
\begin{tabular}{llcccc}
\toprule
\textbf{Model} & \textbf{Metric} & \textbf{seed\_42} & \textbf{seed\_123} & \textbf{seed\_323} & \textbf{mean\std{std}} \\
\midrule
\multirow{3}{*}{BERT}
 & MNLI-m (\%)       & 51.51 & 72.11 & 79.72 & 67.78\std{14.60} \\
 & HANS overall (\%) & 50.14 & 52.90 & 50.94 & 51.33\std{1.42} \\
 & L1 sign-flips     & 4     & 1     & 0     & - \\
\midrule
\multirow{3}{*}{RoBERTa}
 & MNLI-m (\%)       & 87.26 & 87.59 & 87.43 & 87.43\std{0.16} \\
 & HANS overall (\%) & 74.67 & 74.37 & 76.17 & 75.07\std{0.97} \\
 & L1 sign-flips     & 0     & 0     & 0     & - \\
\bottomrule
\end{tabular}
\caption{Per-seed results for canonical L1 DFR (Kirichenko et al., 2023 specification). BERT shows severe seed sensitivity driven by L1 sign-flip instability on SFV features. RoBERTa is fully stable. Sign-flips = number of SFV feature coefficients that change sign relative to the ERM baseline after L1 head retraining.}
\label{tab:l1_dfr}
\end{table}

\section{Full CivilComments Results by Feature Tier}
\label{app:cc_full}

Table~\ref{tab:cc_full} reports CivilComments-WILDS worst-group accuracy for both Tier~C (all 10 SFV features) and Tier~A (causally verified 9-feature subset) across both architectures.
The Tier~A advantage is BERT-specific -- DFR Tier~A gains $+$1.17\,pp over Tier~C (71.84 vs.\ 70.67) by excluding the one feature that does not pass single-target verification on BERT.
RoBERTa shows no consistent Tier~A benefit, suggesting that with stronger representations the full feature set provides equally informative group partitions.

\begin{table}[H]
\centering
\small
\setlength{\tabcolsep}{3pt}
\begin{tabular}{lcccc}
\toprule
 & \multicolumn{2}{c}{\textbf{BERT}} & \multicolumn{2}{c}{\textbf{RoBERTa}} \\
\cmidrule(lr){2-3}\cmidrule(lr){4-5}
\textbf{Method} & \textbf{Tier C} & \textbf{Tier A} & \textbf{Tier C} & \textbf{Tier A} \\
\midrule
ERM           & 58.97\std{1.32} & 58.97\std{1.32} & 55.66\std{0.67} & 55.66\std{0.67} \\
DFR           & 70.67\std{2.79} & \textbf{71.84\std{1.94}} & \textbf{72.12\std{0.26}} & 71.91\std{0.62} \\
DFR-FT        & 65.57\std{0.47} & 67.03\std{2.73} & 71.78\std{1.71} & 71.64\std{1.53} \\
PoE           & 66.03\std{1.26} & 65.96\std{0.99} & 64.88\std{1.58} & 64.41\std{1.44} \\
PoE-IPW-Group & 65.15\std{0.55} & 64.66\std{1.03} & 61.81\std{0.34} & 61.18\std{0.95} \\
SCER          & 67.24\std{0.34} & 67.06\std{0.49} & 67.57\std{0.89} & 67.18\std{0.62} \\
LEACE         & 65.53\std{1.46} & 65.41\std{0.93} & 64.55\std{0.89} & 63.97\std{1.64} \\
\midrule
Kirichenko '23 (lit.) & \multicolumn{2}{c}{70.1\phantom{\std{0.00}}} & \multicolumn{2}{c}{-} \\
\bottomrule
\end{tabular}
\caption{CivilComments-WILDS 16-cell worst-group accuracy (\%) by feature tier. Tier~C = all 10 SFV features, Tier~A = causally verified 9-feature subset (BERT only). \textbf{Bold} = best per architecture. JTT excluded -- BERT ERM error rate ${\approx}3\%$ is too low for upweighting to converge. 3-seed mean\std{std}. Kirichenko '23 uses hand-labeled groups.}
\label{tab:cc_full}
\end{table}

\section{Open-Weight Generator Substitution}
\label{app:open_weight}

\modelname{} uses GPT-4o in its generation roles, which raises a reproducibility concern:
proprietary endpoints are deprecated over time. We therefore re-ran the discovery half of
the pipeline (\textsc{SCGenLLM} $\rightarrow$ deduplication $\rightarrow$ BLV
$\rightarrow$ SFV) on the same 5{,}000-sample MNLI subset with
\textbf{gpt-oss-120b}~\citep{openai2025gptoss}, an Apache-2.0 open-weight model, replacing
GPT-4o in the \textsc{SCGenLLM} and BLV-rewriter roles while keeping Qwen3-32B as
evaluator, so cross-family generator/evaluator decoupling is preserved. The substitution
required no code changes and completed in roughly 32 minutes for about \$3.
Tables~\ref{tab:gptoss_pipeline} and~\ref{tab:gptoss_sfv} report the comparison.

\begin{table}[H]
\centering
\small
\begin{tabular}{lcc}
\toprule
 & \textbf{gpt-oss-120b} & \textbf{GPT-4o} \\
\midrule
Raw \textsc{SCGenLLM} candidates      & 597  & 625 \\
GPT-4o top-$k$ recovered at candidate stage & 9/10 & 10/10 \\
Canonical NLI shortcut classes at SFV   & 3 of 3 & 3 of 3 \\
\bottomrule
\end{tabular}
\caption{Open-weight substitution in the \modelname{} \textbf{discovery} pipeline on the
same 5{,}000-sample MNLI input. GPT-4o recovers 10/10 of its own top-$k$ features by construction,
since Tier~A is defined from the GPT-4o run. The three canonical NLI shortcut classes
(lexical overlap, hypothesis-side negation, length disparity) are recovered by both.
This comparison covers \textsc{SCGenLLM} through SFV only: causal verification,
debiasing, and the discovery baselines of Appendix~\ref{app:discovery_baselines} were
\emph{not} re-run with gpt-oss-120b.}
\label{tab:gptoss_pipeline}
\end{table}

\begin{table}[H]
\centering
\small
\begin{tabular}{lcc}
\toprule
\textbf{Metric} & \textbf{GPT-4o top-10} & \textbf{gpt-oss-120b top-10} \\
\midrule
Mean validation precision  & 0.634\std{0.136} & \textbf{0.698\std{0.144}} \\
Mean validation coverage   & \textbf{0.209\std{0.275}} & 0.183\std{0.238} \\
Mean validation odds ratio & 4.83\std{2.45}   & \textbf{6.86\std{4.14}} \\
Mean SFV score             & 0.468            & \textbf{0.486} \\
\bottomrule
\end{tabular}
\caption{SFV validation statistics for the top-10 features recovered by each generator.
The open-weight run attains higher mean precision ($+$6.4\,pp) and odds ratio ($+$2.03) at
slightly lower coverage, with overall SFV scores within 0.018. An independent Qwen3-32B
similarity judgement over the two top-10 sets returns 8/10, with word overlap,
hypothesis-side negation, length disparity, and contrasting absolutes shared.}
\label{tab:gptoss_sfv}
\end{table}

We therefore recommend gpt-oss-120b as the primary open-weight substitute, with
DeepSeek-V3~\citep{deepseekai2024deepseekv3} as a permissively licensed alternative. The
substitution is not loss-free: the gpt-oss-120b top-10 omits two of the compositional
cross-field features that the GPT-4o run surfaces, which is the axis on which
\modelname{}'s advantage over single-token discovery rests
(Appendix~\ref{app:discovery_baselines}).

\section{Discovery Baseline Comparison}
\label{app:discovery_baselines}

\subsection{NLI}
\label{app:nli_baselines}

Table~\ref{tab:nli_baselines} reports the NLI counterpart of
Table~\ref{tab:discovery_baselines}, using PoE on BERT-base. The three simpler baselines
cluster tightly on the HANS non-entailment lexical-overlap subset (NE-Lex), the split that
directly probes the overlap heuristic, while \modelname{} extends it substantially.
The clustering of LLM-only with PMI to within 0.6\,pp indicates that candidate diversity
alone, without the statistical validation funnel, is not sufficient. PMI's single-token,
single-field formulation cannot express the cross-field compositional features
(length ratios, premise-and-hypothesis conjunctions, contrasting absolutes) that drive the
remaining gap.

\begin{table}[H]
\centering
\small
\begin{tabular}{lccc}
\toprule
\textbf{Method} & \textbf{MNLI-m} & \textbf{HANS} & \textbf{NE-Lex} \\
\midrule
ERM        & 84.61\std{0.19} & 52.41\std{1.65} & 9.53\std{9.78} \\
PMI        & 84.03\std{0.22} & 60.49\std{3.35} & 40.14\std{6.27} \\
PMI+SFV    & 84.13\std{0.28} & 60.46\std{2.75} & 39.60\std{7.68} \\
LLM-only   & 84.31\std{0.25} & 60.47\std{1.78} & 39.86\std{8.28} \\
\midrule
\textbf{\modelname{}} & 83.72\std{0.31} & \textbf{63.84\std{4.08}} & \textbf{54.34\std{18.90}} \\
\bottomrule
\end{tabular}
\caption{Discovery baselines on NLI with BERT-base PoE. MNLI matched accuracy, HANS overall,
and the HANS non-entailment lexical-overlap subset (NE-Lex), all in \%.
\textbf{The three baseline rows were run at seed 42 only} owing to compute budget. The
\modelname{} row is a 3-seed mean. NE-Lex is strongly seed-sensitive on this architecture
(per-seed 50.86 / 74.74 / 37.42), so the baseline comparison should be read as a
single-seed reference point rather than a matched-variance contrast. Per-seed HANS
subcategory values are in Appendix~\ref{app:hans_breakdown}.}
\label{tab:nli_baselines}
\end{table}

\subsection{CivilComments PoE: Cell-Level Decomposition}
\label{app:cc_poe_cells}

Under PoE, \modelname{} trails the lexical baselines on worst-group accuracy
(65.96\std{0.99} against PMI's 69.34\std{1.09}). The 16-cell decomposition shows that the
difference is a systematic trade-off rather than sampling noise: relative to the mean of the
three baselines, \modelname{}'s PoE is more accurate on \emph{all eight} non-toxic
identity cells ($+0.12$ to $+6.10$\,pp, largest on \texttt{black} and \texttt{LGBTQ}) and
less accurate on \emph{all eight} toxic identity cells ($-0.90$ to $-5.86$\,pp). The sign
pattern is identical across every identity axis.

Averaged over all 16 cells the methods are within 0.6\,pp of one another
(\modelname{} 77.64, PMI 78.07, PMI+SFV 78.19, LLM-only 77.97, ERM 77.67), so the WGA
ordering reflects which cell happens to be worst for each method rather than a uniform
accuracy difference. On ERM's worst cell (\texttt{christian}, toxic) \modelname{} reaches
67.30 against PMI's 69.34. Cell-level dispersion is comparable
(\modelname{} 9.32 against 7.62--8.26 for the baselines), and all debiased methods reduce
it substantially from ERM's 14.18.

\section{Qualitative Analysis of Generated Counterfactuals}

\subsection{Counterfactual Generation Example}
\label{app:cf_example}

Table~\ref{tab:cf_example} illustrates a counterfactual generated by our pipeline targeting Feature~150 (rank~1 by SFV score), which captures the spurious heuristic of contrasting absolute quantifiers across premise and hypothesis -- a strong surface-level predictor of \textsc{contradiction} ($\text{OR}=10.01$, precision$=0.83$).
The pipeline successfully removes all four co-occurring spurious features (F150, F130, F91, F25) in a single refinement iteration while preserving the genuine semantic contradiction.

\begin{table}[H]
\centering
\footnotesize
\renewcommand{\arraystretch}{1.3}
\begin{tabular}{p{0.15\linewidth} p{0.75\linewidth}}
\toprule
& \textbf{Text} \\
\midrule

\textbf{Original}\\
\textit{Premise}
& In Nepal, both Hindu and Buddhist temples may take the pagoda form, but \hlc[yellow!50]{all} the Indian-style stone shikara tower temples are Hindu and \hlc[yellow!50]{all} the white dome-like stupas are Buddhist. \\[2pt]

\textit{Hypothesis}
& In Nepal, \hlc[yellow!50]{neither} Hindu \hlc[yellow!50]{nor} Buddhist temples may take the pagoda form because \hlc[yellow!50]{no one} knows this construction style. \\[2pt]

\textit{Label} & \textsc{contradiction} \\

\midrule

\textbf{Counterfactual}\\
\textit{Premise}
& In Nepal, both Hindu and Buddhist temples may take the pagoda form, and Indian-style stone shikara tower temples are Hindu while white dome-like stupas are Buddhist. \\[2pt]

\textit{Hypothesis}
& In Nepal, Hindu and Buddhist temples cannot use the pagoda form because this construction style is unknown. \\[2pt]

\textit{Label} & \textsc{contradiction} \checkmark \\

\midrule

\textbf{Changes}
& Removed \texttt{all} ($\times$2) from premise, replaced \texttt{neither\ldots no one} with \texttt{cannot\ldots unknown} in hypothesis. \\[2pt]

\textbf{Features removed}
& F150 (absolute contrast), F130 (opposite absolutes), F91 (existential + negation), F25 (negation in hypothesis) \\[2pt]

\textbf{Iterations}
& 1 \\

\bottomrule
\end{tabular}
\caption{Counterfactual generated for Feature~150: \textit{``Contrasting extreme words like `always' in premise and `never' in hypothesis''} (SFV rank~1, OR$=10.01$). Highlighted tokens in the original indicate the spurious surface patterns removed by the counterfactual. The semantic contradiction is preserved: the premise confirms that pagoda-form temples {exist} in Nepal, which directly contradicts the hypothesis claim that this construction style is unknown. All four co-occurring spurious features are eliminated in a single generation step.}
\label{tab:cf_example}
\end{table}

\end{document}

%% file: fig_unmask.tex
%
\newcommand{\netedges}[2]{%
  \foreach \a in {3.6,0,-3.6} \foreach \b in {3.6,0,-3.6}
    {\draw[mdl, line width=0.7pt] (#1-4.5,#2+\a) -- (#1,#2+\b);
     \draw[mdl, line width=0.7pt] (#1,#2+\a) -- (#1+4.5,#2+\b);}}
\newcommand{\clffit}[3]{\node[fit={(#2-6.0,#3-5.5) (#2+6.0,#3+5.5)}, inner sep=0] (#1) {};}
\newcommand{\clfA}[3]{%
  \netedges{#2}{#3}
  \foreach \c in {-4.5,0,4.5} \foreach \a in {3.6,0,-3.6}
    {\draw[mdl, line width=1.0pt, fill=shc] (#2+\c,#3+\a) circle (1.25mm);}
  \clffit{#1}{#2}{#3}}
\newcommand{\clfB}[3]{%
  \netedges{#2}{#3}
  \foreach \c in {-4.5,0,4.5} \foreach \a in {3.6,0,-3.6}
    {\draw[mdl, line width=1.0pt, fill=white] (#2+\c,#3+\a) circle (1.25mm);}
  \draw[mdl, line width=1.0pt, fill=shc] (#2,#3+3.6) circle (1.25mm);
  \draw[mdl, line width=1.0pt, fill=shc] (#2+4.5,#3-3.6) circle (1.25mm);
  \clffit{#1}{#2}{#3}}

\newcommand{\docs}[3]{%
  \foreach \o in {1.7,0.85}
    {\draw[cy!50, line width=1.0pt, fill=white, rounded corners=0.7pt]
       (#2-4.0+\o,#3-5.2+\o) rectangle (#2+4.0+\o,#3+5.2+\o);}
  \draw[cy, line width=1.25pt, fill=fy, rounded corners=0.7pt]
       (#2-4.0,#3-5.2) rectangle (#2+4.0,#3+5.2);
  \foreach \k/\w in {3.3/6.2, 1.65/5.0, 0/6.2, -1.65/4.0, -3.3/5.6}
    {\draw[cy!80, line width=0.85pt] (#2-3.0,#3+\k) -- (#2-3.0+\w,#3+\k);}
  \node[fit={(#2-4.3,#3-7.0) (#2+6.0,#3+7.0)}, inner sep=0] (#1) {};}

\newcommand{\sub}{\sffamily\mdseries\footnotesize}
\newcommand{\CAP}[1]{\MakeUppercase{#1}}

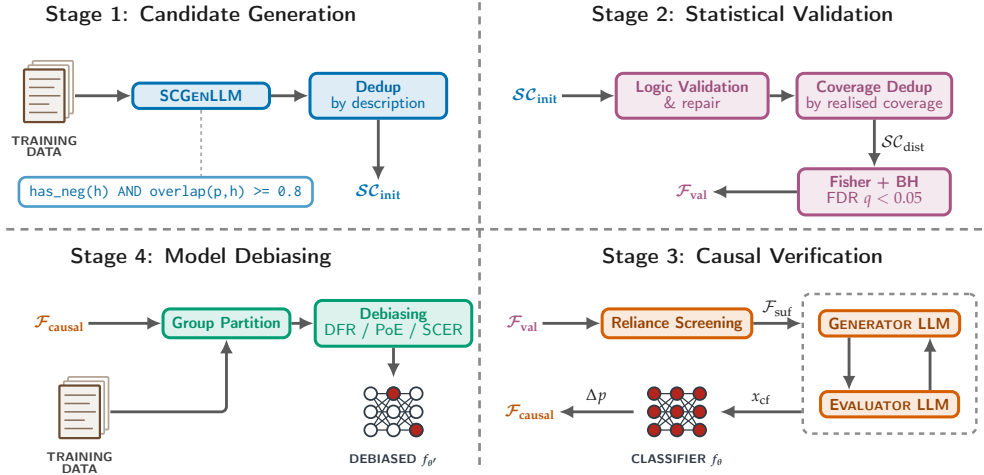
\begin{figure}[t]
\centering
\resizebox{0.93\textwidth}{!}{%
\begin{tikzpicture}[
  x=1mm, y=1mm,
  blk/.style={draw, line width=1.7pt, rounded corners=4pt, align=center,
              inner xsep=5pt, inner ysep=4pt, font=\sffamily\bfseries\small},
  b1/.style={blk, draw=c1, fill=f1, text=c1!92!black},
  b2/.style={blk, draw=c2, fill=f2, text=c2!92!black},
  b3/.style={blk, draw=c3, fill=f3, text=c3!70!black},
  b4/.style={blk, draw=c4, fill=f4, text=c4!92!black},
  tag/.style={inner sep=2pt, font=\bfseries\small\boldmath},
  t1/.style={tag, text=c1},
  t2/.style={tag, text=c2},
  t3/.style={tag, text=c3!88!black},
  code/.style={draw, line width=1.5pt, rounded corners=4pt, align=center,
               inner xsep=6pt, inner ysep=4.5pt, font=\ttfamily\footnotesize},
  cont/.style={draw=ink!50, line width=1.4pt, dash pattern=on 3pt off 3pt,
               rounded corners=4pt},
  fl/.style={-{Latex[length=2.8mm,width=2.6mm]}, line width=1.6pt, ink!72,
             rounded corners=3pt},
  lead/.style={line width=1.1pt, ink!35, dash pattern=on 2pt off 2pt},
  hdr/.style={font=\sffamily\bfseries\large, text=ink},
  glab/.style={font=\sffamily\bfseries\scriptsize, align=center},
  dglab/.style={glab, text=ink!85},
  mglab/.style={glab, text=mdl, font=\sffamily\bfseries\scriptsize\boldmath},
  elab/.style={font=\small, text=ink}
]

\draw[rule, line width=1.5pt, dash pattern=on 4.5pt off 3.5pt] (94,3.5) -- (94,-88);
\draw[rule, line width=1.5pt, dash pattern=on 4.5pt off 3.5pt] (0.5,-41.4) -- (194.0,-41.4);

\node[hdr] at (39,1)    {Stage 1: Candidate Generation};
\node[hdr] at (146,1)   {Stage 2: Statistical Validation};
\node[hdr] at (146,-47) {Stage 3: Causal Verification};
\node[hdr] at (39,-47)  {Stage 4: Model Debiasing};

\docs{cylB}{8}{-15}
\node[dglab] at (8,-25) {\CAP{training}\\[-1.5pt]\CAP{data}};

\node[b1, minimum width=27mm] (gen) at (39,-15) {SCG\scriptsize EN\small LLM};
\node[b1, minimum width=27mm] (dd1) at (74,-15) {Dedup\\[-1pt]{\sub by description}};
\node[code, draw=c1!70, fill=white, text=c1!92!black] (ex) at (32,-34)
     {has\_neg(h) AND overlap(p,h) >= 0.8};
\node[t1] (o1) at (74,-34) {$\mathcal{SC}_{\mathrm{init}}$};

\draw[fl] (cylB.east) -- (gen.west);
\draw[fl] (gen) -- (dd1);
\draw[fl] (dd1) -- (o1);
\draw[lead] (gen.south) -- (gen.south |- ex.north);

\node[t1] (i2) at (105,-15) {$\mathcal{SC}_{\mathrm{init}}$};
\node[b2, minimum width=30mm] (blv) at (136,-15) {Logic Validation\\[-1pt]{\sub \& repair}};
\node[b2, minimum width=30mm] (cov) at (172,-15) {Coverage Dedup\\[-1pt]{\sub by realised coverage}};
\node[b2, minimum width=30mm] (fis) at (172,-34) {Fisher + BH\\[-1pt]{\sub FDR $q<0.05$}};
\node[t2] (o2) at (136,-34) {$\mathcal{F}_{\mathrm{val}}$};

\draw[fl] (i2) -- (blv);
\draw[fl] (blv) -- (cov);
\draw[fl] (cov) -- node[elab,right]{$\mathcal{SC}_{\mathrm{dist}}$} (fis);
\draw[fl] (fis) -- (o2);

\node[t2] (i3) at (102.5,-60) {$\mathcal{F}_{\mathrm{val}}$};
\node[b3] (scr) at (133,-60) {Reliance Screening};
\node[b3] (g)   at (175,-60) {G\scriptsize ENERATOR\small\ LLM};
\node[b3] (e)   at (175,-76) {E\scriptsize VALUATOR\small\ LLM};
\node[t3] (o3)  at (104,-77.5) {$\mathcal{F}_{\mathrm{causal}}$};

\clfA{bm}{133}{-77.5}
\node[mglab] at (133,-87.0) {\CAP{classifier} $f_\theta$};

\begin{scope}[on background layer]
  \node[cont, inner sep=3mm, fit=(g)(e)] (loop) {};
\end{scope}

\draw[fl] (i3) -- (scr);
\draw[fl] (scr.east) -- node[elab,above]{$\mathcal{F}_{\mathrm{suf}}$} (loop.west |- scr);
\draw[fl] ([xshift=-8mm]g.south) -- ([xshift=-8mm]e.north);
\draw[fl] ([xshift=8mm]e.north) -- ([xshift=8mm]g.south);
\draw[fl] (loop.west |- bm) -- node[elab,above]{$x_{\mathrm{cf}}$} ([xshift=2.5mm]bm.east);
\draw[fl] ([xshift=-2.5mm]bm.west) -- node[elab,above]{$\Delta p$} (o3.east);

\node[t3] (i4) at (11,-60) {$\mathcal{F}_{\mathrm{causal}}$};
\node[b4] (grp) at (44,-60) {Group Partition};
\node[b4] (dfr) at (77,-60) {Debiasing\\[-1pt]{\sub DFR / PoE / SCER}};

\docs{cylA}{15}{-77.5}
\node[dglab] at (15,-87.6) {\CAP{training}\\[-1.5pt]\CAP{data}};

\clfB{out}{77}{-77.5}
\node[mglab] at (77,-87.0) {\CAP{debiased} $f_{\theta'}$};

\draw[fl] (i4) -- (grp);
\draw[fl] (grp) -- (dfr);
\draw[fl] (dfr.south) -- ([yshift=1.2mm]out.north);
\draw[fl] (cylA.east) -- (44,-77.5) -- (grp.south);

\end{tikzpicture}}
\caption{\modelname{} makes a classifier's shortcuts nameable, testable, and removable, with no human annotation in the loop. \textbf{Stage 1} prompts an LLM for candidate shortcuts written as \emph{executable} boolean predicates $b(x)$, so each hypothesis can be run on the corpus rather than merely described. \textbf{Stage 2} repairs the predicates, merges those with near-identical realised coverage, and keeps only features whose association with a label survives Fisher's exact test under Benjamini--Hochberg control, yielding $\mathcal{F}_{\mathrm{val}}$ (\S\ref{sec:scdata}). Statistical association is not evidence of use, so \textbf{Stage 3} asks whether $f_\theta$ actually depends on each feature and then edits it out through a generator--evaluator loop. The features whose removal moves the prediction form $\mathcal{F}_{\mathrm{causal}}$ (\S\ref{sec:causal}). Because these features are executable, \textbf{Stage 4} runs them on the training set to obtain group labels for free (\S\ref{sec:dsc}). Filled units mark the shortcut-carrying directions that debiasing removes.}
\label{fig:unmask}
\end{figure}

%% file: cameraready.bbl
\begin{thebibliography}{58}
\providecommand{\natexlab}[1]{#1}
\providecommand{\url}[1]{\texttt{#1}}
\expandafter\ifx\csname urlstyle\endcsname\relax
  \providecommand{\doi}[1]{doi: #1}\else
  \providecommand{\doi}{doi: \begingroup \urlstyle{rm}\Url}\fi

\bibitem[Agarwal et~al.(2020)Agarwal, Shetty, and
  Fritz]{agarwal2020causalvqarevealingreducing}
Vedika Agarwal, Rakshith Shetty, and Mario Fritz.
\newblock Towards causal vqa: Revealing and reducing spurious correlations by
  invariant and covariant semantic editing, 2020.
\newblock URL \url{https://arxiv.org/abs/1912.07538}.

\bibitem[Belrose et~al.(2023)Belrose, Schneider-Joseph, Ravfogel, Cotterell,
  Raff, and Biderman]{belrose2023leace}
Nora Belrose, David Schneider-Joseph, Shauli Ravfogel, Ryan Cotterell, Edward
  Raff, and Stella Biderman.
\newblock {LEACE}: Perfect linear concept erasure in closed form, 2023.
\newblock URL \url{https://arxiv.org/abs/2306.03819}.

\bibitem[Benjamini \& Hochberg(1995)Benjamini and
  Hochberg]{888cd474-50a6-33fd-a789-415b80e67e78}
Yoav Benjamini and Yosef Hochberg.
\newblock Controlling the false discovery rate: A practical and powerful
  approach to multiple testing.
\newblock \emph{Journal of the Royal Statistical Society. Series B
  (Methodological)}, 57\penalty0 (1):\penalty0 289--300, 1995.
\newblock ISSN 00359246.
\newblock URL \url{http://www.jstor.org/stable/2346101}.

\bibitem[Bowman et~al.(2015)Bowman, Angeli, Potts, and
  Manning]{bowman-etal-2015-large}
Samuel~R. Bowman, Gabor Angeli, Christopher Potts, and Christopher~D. Manning.
\newblock A large annotated corpus for learning natural language inference.
\newblock In Llu{\'i}s M{\`a}rquez, Chris Callison-Burch, and Jian Su (eds.),
  \emph{Proceedings of the 2015 Conference on Empirical Methods in Natural
  Language Processing}, pp.\  632--642, Lisbon, Portugal, September 2015.
  Association for Computational Linguistics.
\newblock \doi{10.18653/v1/D15-1075}.
\newblock URL \url{https://aclanthology.org/D15-1075/}.

\bibitem[Clark et~al.(2019)Clark, Yatskar, and
  Zettlemoyer]{clark2019donteasywayout}
Christopher Clark, Mark Yatskar, and Luke Zettlemoyer.
\newblock Don't take the easy way out: Ensemble based methods for avoiding
  known dataset biases, 2019.
\newblock URL \url{https://arxiv.org/abs/1909.03683}.

\bibitem[Creager et~al.(2021)Creager, Jacobsen, and
  Zemel]{creager2021environmentinferenceinvariantlearning}
Elliot Creager, Jörn-Henrik Jacobsen, and Richard Zemel.
\newblock Environment inference for invariant learning, 2021.
\newblock URL \url{https://arxiv.org/abs/2010.07249}.

\bibitem[{DeepSeek-AI}(2024)]{deepseekai2024deepseekv3}
{DeepSeek-AI}.
\newblock Deepseek-v3 technical report.
\newblock \emph{arXiv preprint arXiv:2412.19437}, 2024.
\newblock URL \url{https://arxiv.org/abs/2412.19437}.

\bibitem[Devlin et~al.(2019)Devlin, Chang, Lee, and
  Toutanova]{devlin2019bertpretrainingdeepbidirectional}
Jacob Devlin, Ming-Wei Chang, Kenton Lee, and Kristina Toutanova.
\newblock Bert: Pre-training of deep bidirectional transformers for language
  understanding, 2019.
\newblock URL \url{https://arxiv.org/abs/1810.04805}.

\bibitem[Gardner et~al.(2020)Gardner, Artzi, Basmov, Berant, Bogin, Chen,
  Dasigi, Dua, Elazar, Gottumukkala, Gupta, Hajishirzi, Ilharco, Khashabi, Lin,
  Liu, Liu, Mulcaire, Ning, Singh, Smith, Subramanian, Tsarfaty, Wallace,
  Zhang, and Zhou]{gardner-etal-2020-evaluating}
Matt Gardner, Yoav Artzi, Victoria Basmov, Jonathan Berant, Ben Bogin, Sihao
  Chen, Pradeep Dasigi, Dheeru Dua, Yanai Elazar, Ananth Gottumukkala, Nitish
  Gupta, Hannaneh Hajishirzi, Gabriel Ilharco, Daniel Khashabi, Kevin Lin,
  Jiangming Liu, Nelson~F. Liu, Phoebe Mulcaire, Qiang Ning, Sameer Singh,
  Noah~A. Smith, Sanjay Subramanian, Reut Tsarfaty, Eric Wallace, Ally Zhang,
  and Ben Zhou.
\newblock Evaluating models' local decision boundaries via contrast sets.
\newblock In Trevor Cohn, Yulan He, and Yang Liu (eds.), \emph{Findings of the
  Association for Computational Linguistics: EMNLP 2020}, pp.\  1307--1323,
  Online, November 2020. Association for Computational Linguistics.
\newblock \doi{10.18653/v1/2020.findings-emnlp.117}.
\newblock URL \url{https://aclanthology.org/2020.findings-emnlp.117/}.

\bibitem[Gardner et~al.(2021)Gardner, Merrill, Dodge, Peters, Ross, Singh, and
  Smith]{gardner2021competencyproblemsfindingremoving}
Matt Gardner, William Merrill, Jesse Dodge, Matthew~E. Peters, Alexis Ross,
  Sameer Singh, and Noah~A. Smith.
\newblock Competency problems: On finding and removing artifacts in language
  data, 2021.
\newblock URL \url{https://arxiv.org/abs/2104.08646}.

\bibitem[Geirhos et~al.(2020)Geirhos, Jacobsen, Michaelis, Zemel, Brendel,
  Bethge, and Wichmann]{Geirhos_2020}
Robert Geirhos, Jörn-Henrik Jacobsen, Claudio Michaelis, Richard Zemel,
  Wieland Brendel, Matthias Bethge, and Felix~A. Wichmann.
\newblock Shortcut learning in deep neural networks.
\newblock \emph{Nature Machine Intelligence}, 2\penalty0 (11):\penalty0
  665–673, November 2020.
\newblock ISSN 2522-5839.
\newblock \doi{10.1038/s42256-020-00257-z}.
\newblock URL \url{http://dx.doi.org/10.1038/s42256-020-00257-z}.

\bibitem[Geva et~al.(2019)Geva, Goldberg, and
  Berant]{geva2019modelingtaskannotatorinvestigation}
Mor Geva, Yoav Goldberg, and Jonathan Berant.
\newblock Are we modeling the task or the annotator? an investigation of
  annotator bias in natural language understanding datasets, 2019.
\newblock URL \url{https://arxiv.org/abs/1908.07898}.

\bibitem[Gururangan et~al.(2018)Gururangan, Swayamdipta, Levy, Schwartz,
  Bowman, and Smith]{gururangan2018annotationartifactsnaturallanguage}
Suchin Gururangan, Swabha Swayamdipta, Omer Levy, Roy Schwartz, Samuel~R.
  Bowman, and Noah~A. Smith.
\newblock Annotation artifacts in natural language inference data, 2018.
\newblock URL \url{https://arxiv.org/abs/1803.02324}.

\bibitem[Hosseini et~al.(2025)Hosseini, Nawathe, Moayeri, Balasubramanian, and
  Feizi]{hosseini2025seeingwhatstherespurious}
Parsa Hosseini, Sumit Nawathe, Mazda Moayeri, Sriram Balasubramanian, and
  Soheil Feizi.
\newblock Spurlens: Automatic detection of spurious cues in multimodal llms,
  2025.
\newblock URL \url{https://arxiv.org/abs/2503.08884}.

\bibitem[Jaimes(2025)]{jaimes2025mitigatingspuriouscorrelationsnli}
Christopher~Roman Jaimes.
\newblock Mitigating spurious correlations in {NLI} via {LLM}-synthesized
  counterfactuals and dynamic balanced sampling, 2025.
\newblock URL \url{https://arxiv.org/abs/2512.18462}.

\bibitem[Joshi \& He(2022)Joshi and
  He]{joshi2022investigationineffectivenesscounterfactuallyaugmented}
Nitish Joshi and He~He.
\newblock An investigation of the (in)effectiveness of counterfactually
  augmented data, 2022.
\newblock URL \url{https://arxiv.org/abs/2107.00753}.

\bibitem[Karimi~Mahabadi et~al.(2020)Karimi~Mahabadi, Belinkov, and
  Henderson]{karimi-mahabadi-etal-2020-end}
Rabeeh Karimi~Mahabadi, Yonatan Belinkov, and James Henderson.
\newblock End-to-end bias mitigation by modelling biases in corpora.
\newblock In Dan Jurafsky, Joyce Chai, Natalie Schluter, and Joel Tetreault
  (eds.), \emph{Proceedings of the 58th Annual Meeting of the Association for
  Computational Linguistics}, pp.\  8706--8716, Online, July 2020. Association
  for Computational Linguistics.
\newblock \doi{10.18653/v1/2020.acl-main.769}.
\newblock URL \url{https://aclanthology.org/2020.acl-main.769/}.

\bibitem[Kaushik et~al.(2020)Kaushik, Hovy, and
  Lipton]{kaushik2020learningdifferencemakesdifference}
Divyansh Kaushik, Eduard Hovy, and Zachary~C. Lipton.
\newblock Learning the difference that makes a difference with
  counterfactually-augmented data, 2020.
\newblock URL \url{https://arxiv.org/abs/1909.12434}.

\bibitem[Kirichenko et~al.(2023)Kirichenko, Izmailov, and
  Wilson]{kirichenko2023layerretrainingsufficientrobustness}
Polina Kirichenko, Pavel Izmailov, and Andrew~Gordon Wilson.
\newblock Last layer re-training is sufficient for robustness to spurious
  correlations, 2023.
\newblock URL \url{https://arxiv.org/abs/2204.02937}.

\bibitem[Koh et~al.(2021)Koh, Sagawa, Marklund, Xie, Zhang, Balsubramani, Hu,
  Yasunaga, Phillips, Gao, Lee, David, Stavness, Guo, Earnshaw, Haque, Beery,
  Leskovec, Kundaje, Pierson, Levine, Finn, and Liang]{koh2021wilds}
Pang~Wei Koh, Shiori Sagawa, Henrik Marklund, Sang~Michael Xie, Marvin Zhang,
  Akber Balsubramani, Weihua Hu, Michihiro Yasunaga, Richard~Lanas Phillips,
  Irena Gao, Tony Lee, Etienne David, Ian Stavness, Wei Guo, Berton Earnshaw,
  Imran Haque, Sara~M. Beery, Jure Leskovec, Anshul Kundaje, Emma Pierson,
  Sergey Levine, Chelsea Finn, and Percy Liang.
\newblock {WILDS}: A benchmark of in-the-wild distribution shifts.
\newblock In \emph{Proceedings of the 38th International Conference on Machine
  Learning}, pp.\  5637--5664. PMLR, 2021.
\newblock URL \url{https://arxiv.org/abs/2012.07421}.

\bibitem[Li et~al.(2026)Li, Sun, Huang, Zhong, Jiang, Han, Zhang, Wang, and
  Liu]{li2026preferenceleakagecontaminationproblem}
Dawei Li, Renliang Sun, Yue Huang, Ming Zhong, Bohan Jiang, Jiawei Han,
  Xiangliang Zhang, Wei Wang, and Huan Liu.
\newblock Preference leakage: A contamination problem in llm-as-a-judge, 2026.
\newblock URL \url{https://arxiv.org/abs/2502.01534}.

\bibitem[Liu et~al.(2021)Liu, Haghgoo, Chen, Raghunathan, Koh, Sagawa, Liang,
  and Finn]{liu2021justtraintwiceimproving}
Evan~Zheran Liu, Behzad Haghgoo, Annie~S. Chen, Aditi Raghunathan, Pang~Wei
  Koh, Shiori Sagawa, Percy Liang, and Chelsea Finn.
\newblock Just train twice: Improving group robustness without training group
  information, 2021.
\newblock URL \url{https://arxiv.org/abs/2107.09044}.

\bibitem[Liu et~al.(2022)Liu, Thekinen, Mollaoglu, Tang, Yang, Cheng, Liu, and
  Tang]{liu-etal-2022-toward}
Haochen Liu, Joseph Thekinen, Sinem Mollaoglu, Da~Tang, Ji~Yang, Youlong Cheng,
  Hui Liu, and Jiliang Tang.
\newblock Toward annotator group bias in crowdsourcing.
\newblock In Smaranda Muresan, Preslav Nakov, and Aline Villavicencio (eds.),
  \emph{Proceedings of the 60th Annual Meeting of the Association for
  Computational Linguistics (Volume 1: Long Papers)}, pp.\  1797--1806, Dublin,
  Ireland, May 2022. Association for Computational Linguistics.
\newblock \doi{10.18653/v1/2022.acl-long.126}.
\newblock URL \url{https://aclanthology.org/2022.acl-long.126/}.

\bibitem[Liu et~al.(2019)Liu, Ott, Goyal, Du, Joshi, Chen, Levy, Lewis,
  Zettlemoyer, and Stoyanov]{liu2019robertarobustlyoptimizedbert}
Yinhan Liu, Myle Ott, Naman Goyal, Jingfei Du, Mandar Joshi, Danqi Chen, Omer
  Levy, Mike Lewis, Luke Zettlemoyer, and Veselin Stoyanov.
\newblock Roberta: A robustly optimized bert pretraining approach, 2019.
\newblock URL \url{https://arxiv.org/abs/1907.11692}.

\bibitem[Malik et~al.(2025)Malik, Pyatkin, Land, Morrison, Smith, Hajishirzi,
  and Lambert]{malik2025rewardbench2advancingreward}
Saumya Malik, Valentina Pyatkin, Sander Land, Jacob Morrison, Noah~A. Smith,
  Hannaneh Hajishirzi, and Nathan Lambert.
\newblock Rewardbench 2: Advancing reward model evaluation, 2025.
\newblock URL \url{https://arxiv.org/abs/2506.01937}.

\bibitem[McCoy et~al.(2019)McCoy, Pavlick, and Linzen]{mccoy-etal-2019-right}
R.~Thomas McCoy, Ellie Pavlick, and Tal Linzen.
\newblock Right for the wrong reasons: Diagnosing syntactic heuristics in
  natural language inference.
\newblock In Anna Korhonen, David Traum, and Llu{\'i}s M{\`a}rquez (eds.),
  \emph{Proceedings of the 57th Annual Meeting of the Association for
  Computational Linguistics}, pp.\  3428--3448, Florence, Italy, July 2019.
  Association for Computational Linguistics.
\newblock \doi{10.18653/v1/P19-1334}.
\newblock URL \url{https://aclanthology.org/P19-1334/}.

\bibitem[Menon \& Srivastava(2024)Menon and
  Srivastava]{menon2024discerndecodingsystematicerrors}
Rakesh~R. Menon and Shashank Srivastava.
\newblock Discern: Decoding systematic errors in natural language for text
  classifiers, 2024.
\newblock URL \url{https://arxiv.org/abs/2410.22239}.

\bibitem[Mu et~al.(2024)Mu, Helyar, Heidecke, Achiam, Vallone, Kivlichan, Lin,
  Beutel, Schulman, and Weng]{mu2024rulebasedrewardslanguage}
Tong Mu, Alec Helyar, Johannes Heidecke, Joshua Achiam, Andrea Vallone, Ian
  Kivlichan, Molly Lin, Alex Beutel, John Schulman, and Lilian Weng.
\newblock Rule based rewards for language model safety, 2024.
\newblock URL \url{https://arxiv.org/abs/2411.01111}.

\bibitem[Müllner(2011)]{mullner2011modernhierarchicalagglomerativeclustering}
Daniel Müllner.
\newblock Modern hierarchical, agglomerative clustering algorithms, 2011.
\newblock URL \url{https://arxiv.org/abs/1109.2378}.

\bibitem[Nam et~al.(2020)Nam, Cha, Ahn, Lee, and
  Shin]{nam2020learningfailuretrainingdebiased}
Junhyun Nam, Hyuntak Cha, Sungsoo Ahn, Jaeho Lee, and Jinwoo Shin.
\newblock Learning from failure: Training debiased classifier from biased
  classifier, 2020.
\newblock URL \url{https://arxiv.org/abs/2007.02561}.

\bibitem[Nie et~al.(2020{\natexlab{a}})Nie, Williams, Dinan, Bansal, Weston,
  and Kiela]{nie-etal-2020-adversarial}
Yixin Nie, Adina Williams, Emily Dinan, Mohit Bansal, Jason Weston, and Douwe
  Kiela.
\newblock Adversarial {NLI}: A new benchmark for natural language
  understanding.
\newblock In Dan Jurafsky, Joyce Chai, Natalie Schluter, and Joel Tetreault
  (eds.), \emph{Proceedings of the 58th Annual Meeting of the Association for
  Computational Linguistics}, pp.\  4885--4901, Online, July
  2020{\natexlab{a}}. Association for Computational Linguistics.
\newblock \doi{10.18653/v1/2020.acl-main.441}.
\newblock URL \url{https://aclanthology.org/2020.acl-main.441/}.

\bibitem[Nie et~al.(2020{\natexlab{b}})Nie, Zhou, and Bansal]{nie2020chaosnli}
Yixin Nie, Xiang Zhou, and Mohit Bansal.
\newblock What can we learn from collective human opinions on natural language
  inference data?
\newblock In \emph{Proceedings of the 2020 Conference on Empirical Methods in
  Natural Language Processing (EMNLP)}, pp.\  9131--9143, Online, November
  2020{\natexlab{b}}. Association for Computational Linguistics.
\newblock \doi{10.18653/v1/2020.emnlp-main.734}.
\newblock URL \url{https://aclanthology.org/2020.emnlp-main.734/}.

\bibitem[{OpenAI}(2025)]{openai2025gptoss}
{OpenAI}.
\newblock gpt-oss-120b \& gpt-oss-20b model card.
\newblock \emph{arXiv preprint arXiv:2508.10925}, 2025.
\newblock URL \url{https://arxiv.org/abs/2508.10925}.

\bibitem[{OpenAI} et~al.(2024){OpenAI}, Hurst, Lerer,
  et~al.]{openai2024gpt4ocard}
{OpenAI}, Aaron Hurst, Adam Lerer, et~al.
\newblock Gpt-4o system card, 2024.
\newblock URL \url{https://arxiv.org/abs/2410.21276}.

\bibitem[Panickssery et~al.(2024)Panickssery, Bowman, and
  Feng]{panickssery2024llmevaluatorsrecognizefavor}
Arjun Panickssery, Samuel~R. Bowman, and Shi Feng.
\newblock Llm evaluators recognize and favor their own generations, 2024.
\newblock URL \url{https://arxiv.org/abs/2404.13076}.

\bibitem[Park et~al.(2026)Park, Kim, Lee, Yoo, and
  Song]{park2026spuriouscorrelationawareembeddingregularization}
Subeen Park, Joowang Kim, Hakyung Lee, Sunjae Yoo, and Kyungwoo Song.
\newblock Spurious correlation-aware embedding regularization for worst-group
  robustness, 2026.
\newblock URL \url{https://arxiv.org/abs/2511.04401}.

\bibitem[Poliak et~al.(2018)Poliak, Naradowsky, Haldar, Rudinger, and
  Van~Durme]{poliak-etal-2018-hypothesis}
Adam Poliak, Jason Naradowsky, Aparajita Haldar, Rachel Rudinger, and Benjamin
  Van~Durme.
\newblock Hypothesis only baselines in natural language inference.
\newblock In Malvina Nissim, Jonathan Berant, and Alessandro Lenci (eds.),
  \emph{Proceedings of the Seventh Joint Conference on Lexical and
  Computational Semantics}, pp.\  180--191, New Orleans, Louisiana, June 2018.
  Association for Computational Linguistics.
\newblock \doi{10.18653/v1/S18-2023}.
\newblock URL \url{https://aclanthology.org/S18-2023/}.

\bibitem[Ratner et~al.(2017)Ratner, Bach, Ehrenberg, Fries, Wu, and
  Ré]{Ratner_2017}
Alexander Ratner, Stephen~H. Bach, Henry Ehrenberg, Jason Fries, Sen Wu, and
  Christopher Ré.
\newblock Snorkel: rapid training data creation with weak supervision.
\newblock \emph{Proceedings of the VLDB Endowment}, 11\penalty0 (3):\penalty0
  269–282, November 2017.
\newblock ISSN 2150-8097.
\newblock \doi{10.14778/3157794.3157797}.
\newblock URL \url{http://dx.doi.org/10.14778/3157794.3157797}.

\bibitem[Ratner et~al.(2018)Ratner, Hancock, Dunnmon, Sala, Pandey, and
  Ré]{ratner2018trainingcomplexmodelsmultitask}
Alexander Ratner, Braden Hancock, Jared Dunnmon, Frederic Sala, Shreyash
  Pandey, and Christopher Ré.
\newblock Training complex models with multi-task weak supervision, 2018.
\newblock URL \url{https://arxiv.org/abs/1810.02840}.

\bibitem[Ribeiro et~al.(2020)Ribeiro, Wu, Guestrin, and
  Singh]{ribeiro-etal-2020-beyond}
Marco~Tulio Ribeiro, Tongshuang Wu, Carlos Guestrin, and Sameer Singh.
\newblock Beyond accuracy: Behavioral testing of {NLP} models with
  {C}heck{L}ist.
\newblock In Dan Jurafsky, Joyce Chai, Natalie Schluter, and Joel Tetreault
  (eds.), \emph{Proceedings of the 58th Annual Meeting of the Association for
  Computational Linguistics}, pp.\  4902--4912, Online, July 2020. Association
  for Computational Linguistics.
\newblock \doi{10.18653/v1/2020.acl-main.442}.
\newblock URL \url{https://aclanthology.org/2020.acl-main.442/}.

\bibitem[Sagawa et~al.(2020)Sagawa, Koh, Hashimoto, and
  Liang]{sagawa2020distributionallyrobustneuralnetworks}
Shiori Sagawa, Pang~Wei Koh, Tatsunori~B. Hashimoto, and Percy Liang.
\newblock Distributionally robust neural networks for group shifts: On the
  importance of regularization for worst-case generalization, 2020.
\newblock URL \url{https://arxiv.org/abs/1911.08731}.

\bibitem[Sanh et~al.(2020)Sanh, Wolf, Belinkov, and
  Rush]{sanh2020learningothersmistakesavoiding}
Victor Sanh, Thomas Wolf, Yonatan Belinkov, and Alexander~M. Rush.
\newblock Learning from others' mistakes: Avoiding dataset biases without
  modeling them, 2020.
\newblock URL \url{https://arxiv.org/abs/2012.01300}.

\bibitem[Schuster et~al.(2019)Schuster, Shah, Yeo, Filizzola, Santus, and
  Barzilay]{schuster2019debiasingfactverificationmodels}
Tal Schuster, Darsh~J Shah, Yun Jie~Serene Yeo, Daniel Filizzola, Enrico
  Santus, and Regina Barzilay.
\newblock Towards debiasing fact verification models, 2019.
\newblock URL \url{https://arxiv.org/abs/1908.05267}.

\bibitem[Sharma et~al.(2025)Sharma, Tong, Korbak, Duvenaud, Askell, Bowman,
  Cheng, Durmus, Hatfield-Dodds, Johnston, Kravec, Maxwell, McCandlish,
  Ndousse, Rausch, Schiefer, Yan, Zhang, and
  Perez]{sharma2025understandingsycophancylanguagemodels}
Mrinank Sharma, Meg Tong, Tomasz Korbak, David Duvenaud, Amanda Askell,
  Samuel~R. Bowman, Newton Cheng, Esin Durmus, Zac Hatfield-Dodds, Scott~R.
  Johnston, Shauna Kravec, Timothy Maxwell, Sam McCandlish, Kamal Ndousse,
  Oliver Rausch, Nicholas Schiefer, Da~Yan, Miranda Zhang, and Ethan Perez.
\newblock Towards understanding sycophancy in language models, 2025.
\newblock URL \url{https://arxiv.org/abs/2310.13548}.

\bibitem[Singhal et~al.(2024)Singhal, Goyal, Xu, and
  Durrett]{singhal2024longwaygoinvestigating}
Prasann Singhal, Tanya Goyal, Jiacheng Xu, and Greg Durrett.
\newblock A long way to go: Investigating length correlations in rlhf, 2024.
\newblock URL \url{https://arxiv.org/abs/2310.03716}.

\bibitem[Sohoni et~al.(2022)Sohoni, Dunnmon, Angus, Gu, and
  Ré]{sohoni2022subclassleftbehindfinegrained}
Nimit~S. Sohoni, Jared~A. Dunnmon, Geoffrey Angus, Albert Gu, and Christopher
  Ré.
\newblock No subclass left behind: Fine-grained robustness in coarse-grained
  classification problems, 2022.
\newblock URL \url{https://arxiv.org/abs/2011.12945}.

\bibitem[Srikanth \& Rudinger(2022)Srikanth and
  Rudinger]{srikanth-rudinger-2022-partial}
Neha Srikanth and Rachel Rudinger.
\newblock Partial-input baselines show that {NLI} models can ignore context,
  but they don{'}t.
\newblock In Marine Carpuat, Marie-Catherine de~Marneffe, and Ivan~Vladimir
  Meza~Ruiz (eds.), \emph{Proceedings of the 2022 Conference of the North
  American Chapter of the Association for Computational Linguistics: Human
  Language Technologies}, pp.\  4753--4763, Seattle, United States, July 2022.
  Association for Computational Linguistics.
\newblock \doi{10.18653/v1/2022.naacl-main.350}.
\newblock URL \url{https://aclanthology.org/2022.naacl-main.350/}.

\bibitem[Sundararajan et~al.(2017)Sundararajan, Taly, and
  Yan]{sundararajan2017axiomaticattributiondeepnetworks}
Mukund Sundararajan, Ankur Taly, and Qiqi Yan.
\newblock Axiomatic attribution for deep networks, 2017.
\newblock URL \url{https://arxiv.org/abs/1703.01365}.

\bibitem[Upton(2018)]{10.2307/2982890}
Graham J.~G. Upton.
\newblock Fisher’s exact test.
\newblock \emph{Journal of the Royal Statistical Society Series A: Statistics
  in Society}, 155\penalty0 (3):\penalty0 395--402, 12 2018.
\newblock ISSN 0964-1998.
\newblock \doi{10.2307/2982890}.
\newblock URL \url{https://doi.org/10.2307/2982890}.

\bibitem[Wang et~al.(2022)Wang, Sridhar, Yang, and
  Wang]{wang-etal-2022-identifying}
Tianlu Wang, Rohit Sridhar, Diyi Yang, and Xuezhi Wang.
\newblock Identifying and mitigating spurious correlations for improving
  robustness in {NLP} models.
\newblock In Marine Carpuat, Marie-Catherine de~Marneffe, and Ivan~Vladimir
  Meza~Ruiz (eds.), \emph{Findings of the Association for Computational
  Linguistics: NAACL 2022}, pp.\  1719--1729, Seattle, United States, July
  2022. Association for Computational Linguistics.
\newblock \doi{10.18653/v1/2022.findings-naacl.130}.
\newblock URL \url{https://aclanthology.org/2022.findings-naacl.130/}.

\bibitem[Wang \& Culotta(2020)Wang and Culotta]{wang-culotta-2020-identifying}
Zhao Wang and Aron Culotta.
\newblock Identifying spurious correlations for robust text classification.
\newblock In \emph{Findings of the Association for Computational Linguistics:
  EMNLP 2020}, pp.\  3431--3440. Association for Computational Linguistics,
  2020.
\newblock URL \url{https://aclanthology.org/2020.findings-emnlp.308/}.

\bibitem[Williams et~al.(2018)Williams, Nangia, and
  Bowman]{williams2018broadcoveragechallengecorpussentence}
Adina Williams, Nikita Nangia, and Samuel~R. Bowman.
\newblock A broad-coverage challenge corpus for sentence understanding through
  inference, 2018.
\newblock URL \url{https://arxiv.org/abs/1704.05426}.

\bibitem[Wu et~al.(2022)Wu, Gardner, Stenetorp, and
  Dasigi]{wu2022generatingdatamitigatespurious}
Yuxiang Wu, Matt Gardner, Pontus Stenetorp, and Pradeep Dasigi.
\newblock Generating data to mitigate spurious correlations in natural language
  inference datasets, 2022.
\newblock URL \url{https://arxiv.org/abs/2203.12942}.

\bibitem[Yang et~al.(2025)Yang, Li, Yang, Zhang, Hui, Zheng, Yu, Gao, Huang,
  Lv, Zheng, Liu, Zhou, Huang, Hu, Ge, Wei, Lin, Tang, Yang, Tu, Zhang, Yang,
  Yang, Zhou, Zhou, Lin, Dang, Bao, Yang, Yu, Deng, Li, Xue, Li, Zhang, Wang,
  Zhu, Men, Gao, Liu, Luo, Li, Tang, Yin, Ren, Wang, Zhang, Ren, Fan, Su,
  Zhang, Zhang, Wan, Liu, Wang, Cui, Zhang, Zhou, and
  Qiu]{yang2025qwen3technicalreport}
An~Yang, Anfeng Li, Baosong Yang, Beichen Zhang, Binyuan Hui, Bo~Zheng, Bowen
  Yu, Chang Gao, Chengen Huang, Chenxu Lv, Chujie Zheng, Dayiheng Liu, Fan
  Zhou, Fei Huang, Feng Hu, Hao Ge, Haoran Wei, Huan Lin, Jialong Tang, Jian
  Yang, Jianhong Tu, Jianwei Zhang, Jianxin Yang, Jiaxi Yang, Jing Zhou,
  Jingren Zhou, Junyang Lin, Kai Dang, Keqin Bao, Kexin Yang, Le~Yu, Lianghao
  Deng, Mei Li, Mingfeng Xue, Mingze Li, Pei Zhang, Peng Wang, Qin Zhu, Rui
  Men, Ruize Gao, Shixuan Liu, Shuang Luo, Tianhao Li, Tianyi Tang, Wenbiao
  Yin, Xingzhang Ren, Xinyu Wang, Xinyu Zhang, Xuancheng Ren, Yang Fan, Yang
  Su, Yichang Zhang, Yinger Zhang, Yu~Wan, Yuqiong Liu, Zekun Wang, Zeyu Cui,
  Zhenru Zhang, Zhipeng Zhou, and Zihan Qiu.
\newblock Qwen3 technical report, 2025.
\newblock URL \url{https://arxiv.org/abs/2505.09388}.

\bibitem[Yang et~al.(2024)Yang, won Hwang, and
  So]{yang2024relationbasedcounterfactualdataaugmentation}
Heerin Yang, Sseung won Hwang, and Jungmin So.
\newblock Relation-based counterfactual data augmentation and contrastive
  learning for robustifying natural language inference models, 2024.
\newblock URL \url{https://arxiv.org/abs/2410.20710}.

\bibitem[Zhang et~al.(2025)Zhang, Xiong, Chen, Zhou, Huang, and
  Zhang]{zhang2025listsemojisformatbias}
Xuanchang Zhang, Wei Xiong, Lichang Chen, Tianyi Zhou, Heng Huang, and Tong
  Zhang.
\newblock From lists to emojis: How format bias affects model alignment, 2025.
\newblock URL \url{https://arxiv.org/abs/2409.11704}.

\bibitem[Zheng et~al.(2025)Zheng, Ye, and
  Zhang]{zheng2025shortcutprobeprobingpredictionshortcuts}
Guangtao Zheng, Wenqian Ye, and Aidong Zhang.
\newblock Shortcutprobe: Probing prediction shortcuts for learning robust
  models, 2025.
\newblock URL \url{https://arxiv.org/abs/2505.13910}.

\bibitem[Zhou et~al.(2024)Zhou, Xu, Liu, An, Ai, and
  Huang]{zhou2024explorespuriouscorrelationsconcept}
Yuhang Zhou, Paiheng Xu, Xiaoyu Liu, Bang An, Wei Ai, and Furong Huang.
\newblock Explore spurious correlations at the concept level in language models
  for text classification, 2024.
\newblock URL \url{https://arxiv.org/abs/2311.08648}.

\end{thebibliography}
